\documentclass[preprint,12pt]{elsarticle}
\usepackage{multirow}
\usepackage{graphicx}
\usepackage{amssymb}
\usepackage{subfigmat}
\usepackage{subfig}
\usepackage{graphicx}
\usepackage{multirow}
\usepackage{amssymb}
\usepackage{here}
\usepackage{appendix}
\usepackage{url}
\usepackage[top=2cm, bottom=2cm, left=1.5cm, right=1.5cm]{geometry}

\usepackage{setspace} 
\journal{Computers and Electronics in Agriculture}

\begin{document}

\begin{frontmatter}

\title{Investigation to answer three key questions concerning plant pest identification and development of a practical identification framework}

\author[inst1]{Ryosuke Wayama}
\author[inst1]{Yuki Sasaki}
\affiliation[inst1]{organization={Northern System Service Co., Ltd.},
            addressline={4-3-5 Motomiya}, 
            city={Morioka},
            postcode={020-0866}, 
            state={Iwate},
            country={Japan}}
\author[inst2]{Satoshi Kagiwada}
\affiliation[inst2]{organization={Department of Clinical Plant Science, Faculty of Bio Science and Applied Chemistry, Hosei University},
            addressline={3-7-2 Kajino}, 
            city={Koganei},
            postcode={184-8584}, 
            state={Tokyo},
            country={Japan}}

\author[inst3]{Nobusuke Iwasaki}
\affiliation[inst3]{organization={Institute for Agro-Environmental Sciences, NARO},
            addressline={3-1-1, Kannondai}, 
            city={Tsukuba},
            postcode={305-8604}, 
            state={Ibaraki},
            country={Japan}}

\author[inst4]{Hitoshi Iyatomi}
\affiliation[inst4]{organization={Department of Applied Informatics, Faculty of Science and Engineering, Hosei University},
            addressline={3-7-2 Kajino}, 
            city={Koganei},
            postcode={184-8584}, 
            state={Tokyo},
            country={Japan}}

\begin{abstract}
The development of practical and robust automated diagnostic systems for identifying plant pests is crucial for efficient agricultural production.
In this paper, we first investigate three key research questions (RQs) that have not been addressed thus far in the field of image-based plant pest identification.
Based on the knowledge gained, we then develop an accurate, robust, and fast plant pest identification framework using 334K images comprising 78 combinations of four plant portions (the leaf front, leaf back, fruit, and flower of cucumber, tomato, strawberry, and eggplant) and 20 pest species captured at 27 farms.
The results reveal the following. 
(1) For an appropriate evaluation of the model, the test data should not include images of the field from which the training images were collected, or other considerations to increase the diversity of the test set should be taken into account.
(2) Pre-extraction of ROIs, such as leaves and fruits, helps to improve identification accuracy.
(3) Integration of closely related species using the same control methods and cross-crop training methods for the same pests, are effective.
Our two-stage plant pest identification framework, enabling ROI detection and convolutional neural network (CNN)-based identification, achieved a highly practical performance of 91.0\% and 88.5\% in mean accuracy and macro F1 score, respectively, for 12,223 instances of test data of 21 classes collected from unseen fields, where 25 classes of images from 318,971 samples were used for training; the average identification time was 476 ms/image.
\end{abstract}

\begin{highlights}
\item Development of an accurate and practical framework for plant pest identification.
\item Evaluation of the model requires images, such as those collected elsewhere.
\item Foreground extraction significantly improves pest discrimination performance.
\item Class integration and cross-crop training boosts model performance.
\end{highlights}

\begin{keyword}
pest identification \sep pest discrimination \sep automated plant diagnosis \sep machine learning \sep ROI detection
\end{keyword}

\end{frontmatter}


\section{Introduction}
\label{sec:intro}
Crop pests are a major threat to agricultural production.
According to the Food and Agriculture Organization of the United Nations (FAO) \cite{FAO2019}, 
20\%--40\% of global agricultural production is lost due to pest damage. 
Thus, identifying pest species is essential for the effective management of agricultural pests, but it is an expensive and time-consuming process.
To address this issue, with the development of image analysis technology using deep learning, particularly represented by convolutional neural networks (CNNs), research has been conducted on techniques of automated pest identification from images of pest-damaged plants 
\cite{Liu2016,Fuentes2017,Selvaraj2019,Li2019,Ren2019,Tetila2019,Wang2020, Lin2020, Kusrini2020, Tassis2021, Guo2021, Rong2022, Wu2019, Bollis2020, Kong2022, Wang2021}.
However, pest identification differs from general image classification and object recognition tasks, as it relies on the insect’s body shape, as well as feeding and sucking marks.
Because insect bodies vary in size depending on the pest species, identifying millimeter- or sub-millimeter-sized pests is challenging, as they are extremely small on the image (e.g.,  approximately 10 pixels per square or less in a 1024 $\times$ 1024 image).
Pest damage scars appear as deformations and discolorations of the injured area of the plant, but the initial damage is often indistinguishable from healthy leaves or fruit. 
Furthermore, there are significant variations in visual characteristics within a single pest species, which result from differences in their growth stages and patterns of damage.
Thus, the task of plant pest identification is characterized by small feature differences between classes and high intra-class diversity, thereby making it a challenging and so-called fine-grained task.
This is also true for plant disease diagnosis, which presents similar difficulties because lesions are often ambiguous and because intra-class diversity is often greater than inter-class differences 
\cite{Lu2021, Liu2021}. 

Figure 1 presents example images of the pests included in this study: in (b) and (c), the insect body and the white-spotted (scabrous) sucking marks on the leaves, respectively, are evidence of pest damage. 
Further, Figures 1 (d) and (e) represent the same pest species, but the morphologies differ completely. ((d): remnants of laid eggs, (e): old larvae). 
A partial enlargement (f) is excluded from this study, where the insect body was magnified by macro photography.
%
%
This study focused on analyzing images of leaves and other areas showing signs of pest damage, taken with a standard camera, rather than highly magnified images of insect bodies.
This choice was motivated by several factors: (1) insect bodies are often not visible, necessitating reliance on feeding scars and other clues for pest damage diagnosis, (2) requiring specialized equipment can be inconvenient for users, and (3) simultaneous diagnosis of both disease and pest damage is preferred.
Given the commonality of disease diagnosis using smartphones, pest damage detection using similar devices is ideal.
%
%
%
%
Although numerous pest identification studies using deep learning techniques report highly numerically accurate results (i.e., 90\%--99\% in average accuracy or F1), they are often based on small datasets and are often unreliable, including problems with the evaluation methods described below.
%

\begin{figure}[t]
\begin{subfigmatrix}{3}
 \subfigure[Healthy]{\includegraphics[]{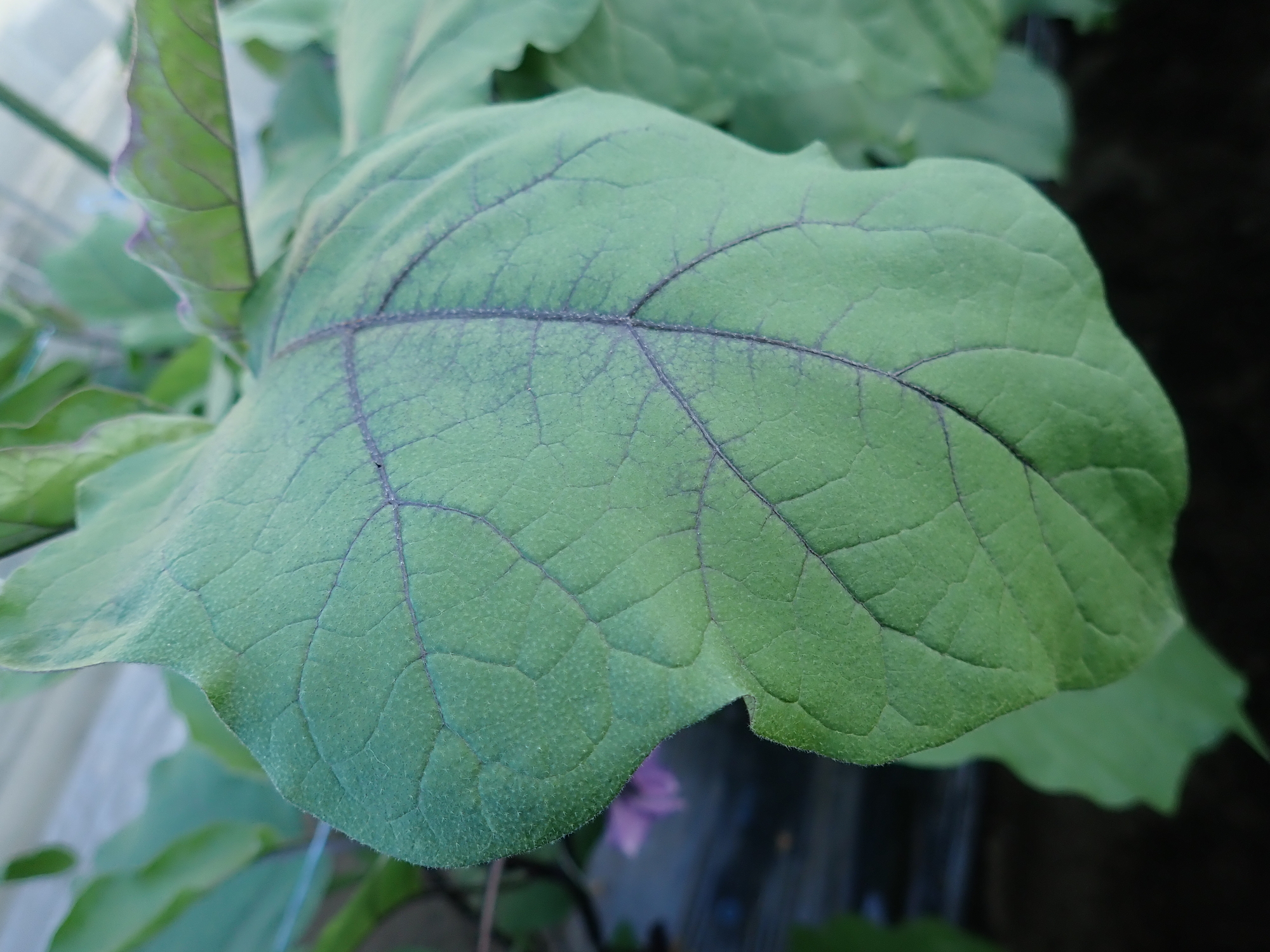}}
 \subfigure[Tobacco whitefly]{\includegraphics[]{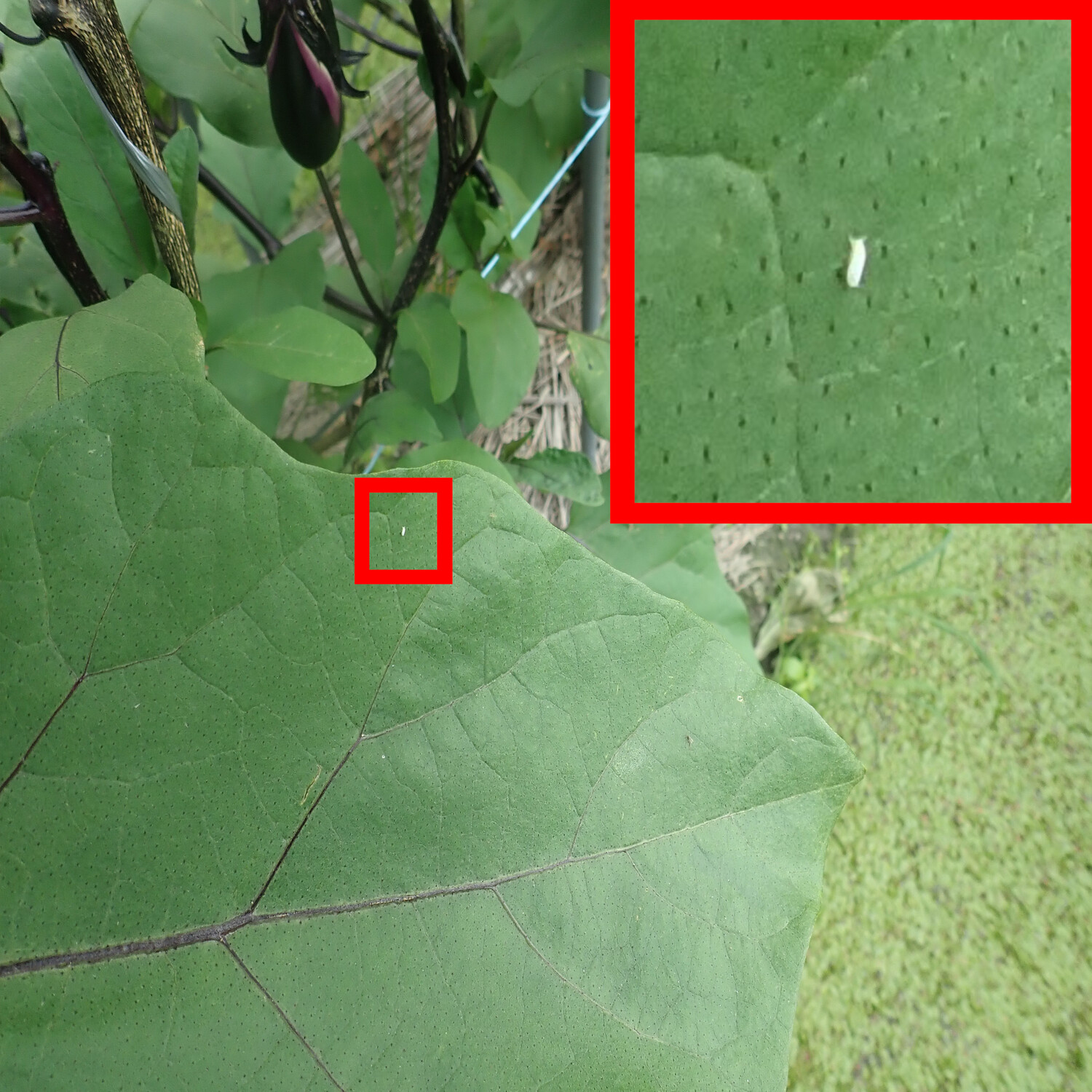}}
 \subfigure[Kanzawa spider mite]{\includegraphics[]{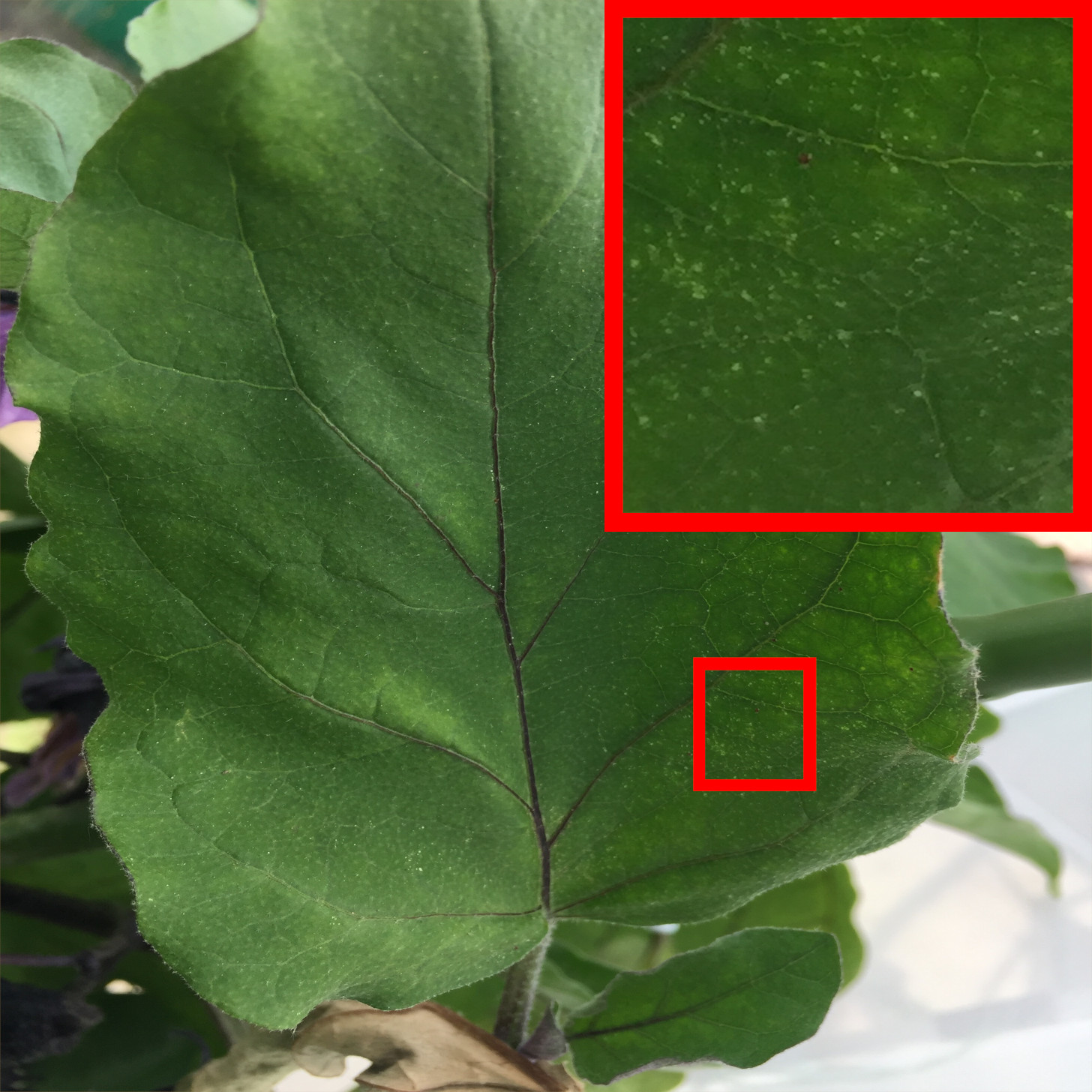}}
 \subfigure[Tobacco cutworm]{\includegraphics[]{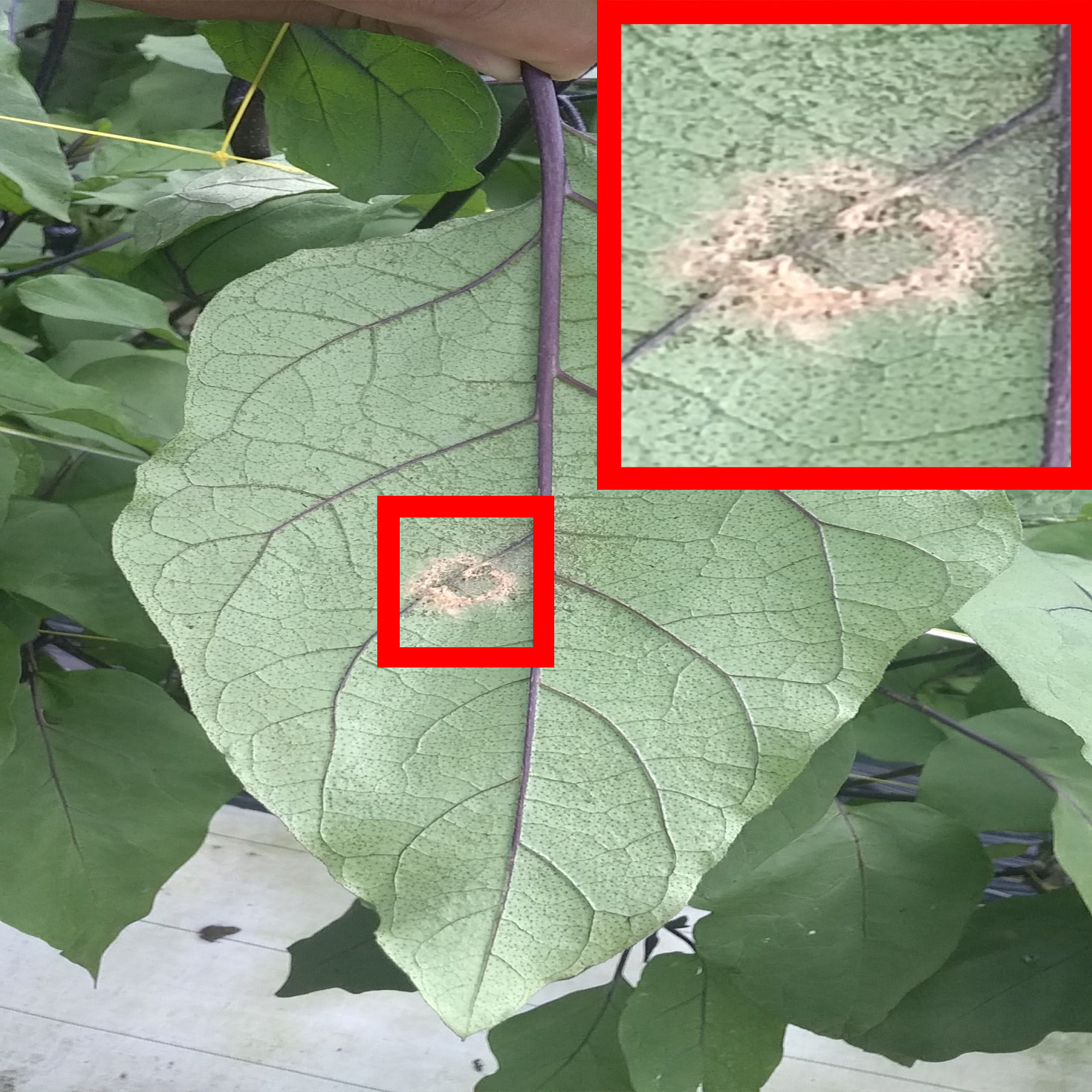}}
 \subfigure[Tobacco cutworm]{\includegraphics[]{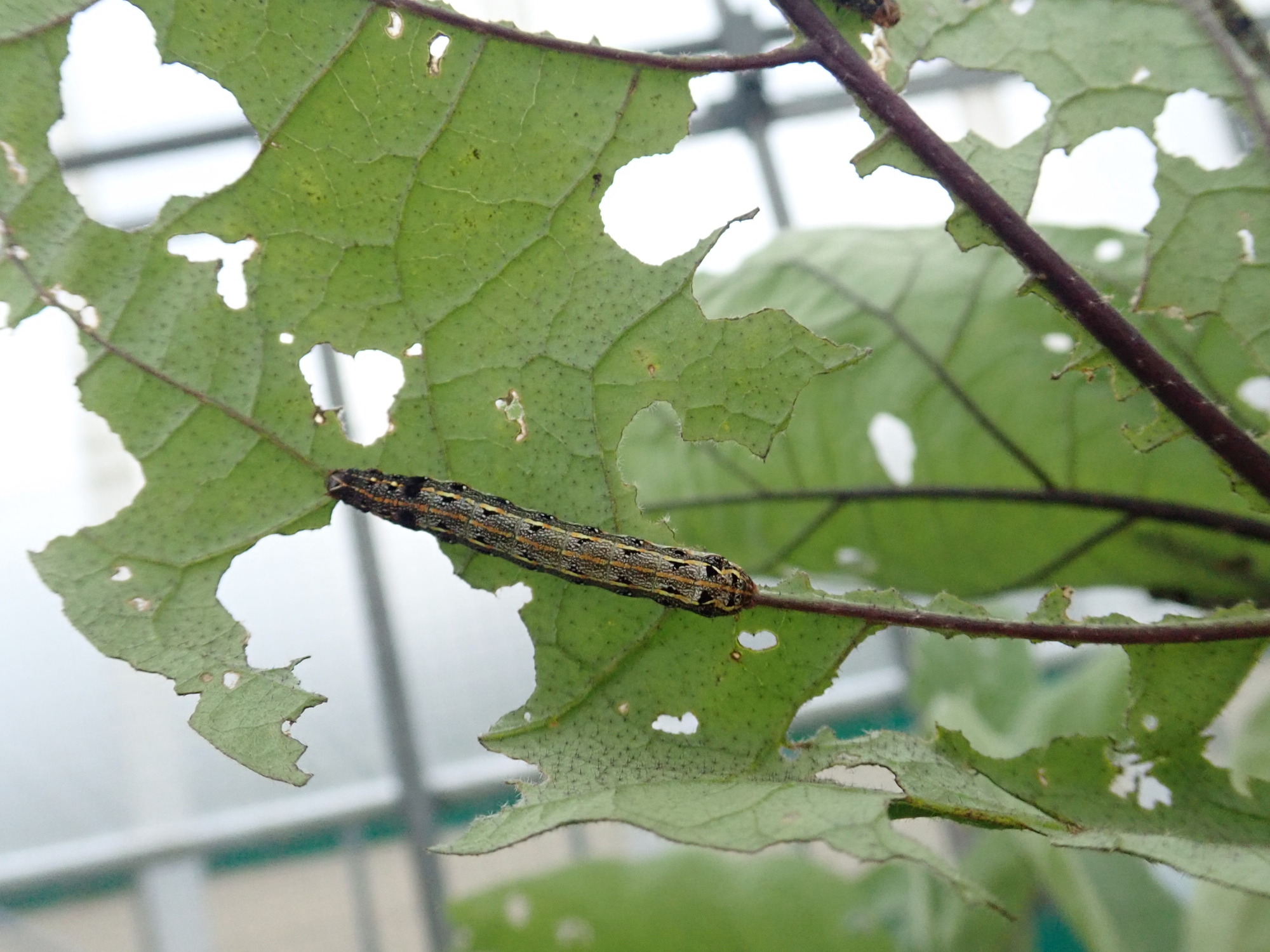}}
 \subfigure[Excluded]{\includegraphics[]{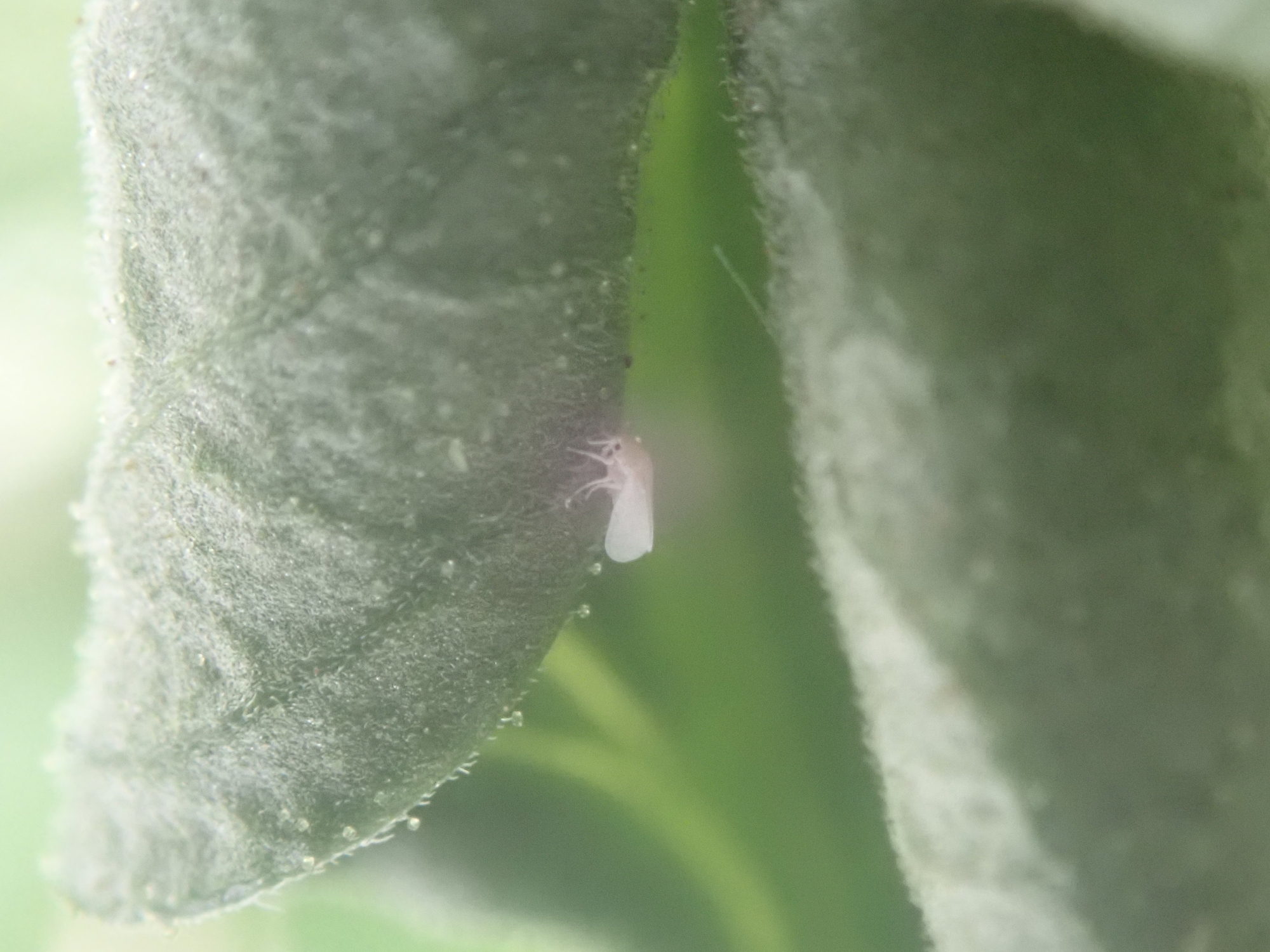}}
\end{subfigmatrix}
\caption{Example of images used and not used in this study. (a)--(e): involved in this study (f): excluded from this study where the insect body was magnified by macro photography. 
In these examples, the insect bodies were approximately 5 pixels square with an image size of 1024$\times$1024 image size, and in some cases the smallest spider mite detected in this experiment was 3--4 pixels square.
In cases where the pest's body size was too small, the model used feeding scars and other cues to identify the pest species.}
\label{fig:fig1}
\end{figure}

Conversely, there has been some research in recent years on the construction of large pest datasets and automated pest diagnosis systems based on these datasets.
%
For example, Wu et al. \cite{Wu2019} constructed and published the IP102 dataset consisting of 75K insect images in 102 species, collected from images and videos of public datasets on the Internet.
Their best model, evaluated by K-nearest neighbor (k = 5) on the representations acquired by ResNet50, achieved F1 = 40.5.
Bollis et al. \cite{Bollis2020} constructed a citrus pest benchmark (CPB) dataset consisting of 10.8K images of insect bodies of seven classes and taken at 60x magnification.
Their original CNN model achieved 91.8\% accuracy on this dataset, and they reported F1 = 59.6 on the IP102 dataset with EfficientNet-B0, a cutting-edge CNN.
Kong et al. \cite{Kong2022} constructed and published the CropDP-181 dataset consisting of 124K images of 181 classes (134 pests and 47 diseases) by extracting insect body images from two public datasets (iNaturalist, AIChallenger) and integrating them with the IP102 dataset.
Their feature-enhanced attention neural network (Fe-Net) reported a top-1 accuracy of 85.3\%.
However, even in analyses based on such large-scale data, only insect bodies were the focus of their studies; damage scars---which are important evidence for discriminating unphotographable insect pests---were not included in the analysis.
Partially because of this, Wang et al. \cite{Wang2021} noted that most of the IP102 dataset is dominated by diseases that are not agriculturally harmful and constructed the AgriPest dataset consisting of 14 pests, 4 crops, and 49.7K images (with 264.6K annotated insect body parts).
Unlike the IP102 and CPB, this dataset features images shot from a consistent angle, where insects occupy approxmately 0.16\% of the image area on average, with a maximum representation of 3\%.
They evaluated six different object detection models, and their Cascade R-CNN achieved 70.83 mAP.

The provision of these large datasets is a highly significant contribution to research in this area, and the performance of discriminators based on these datasets has been reported to be good.
Conversely, most studies, including those employing large datasets, raise the important question of whether the models have been adequately evaluated.
In this paper, to achieve practical and automatic plant pest identification, we first investigate three issues.

First, a serious problem with existing plant pest identification studies is that the models may not be adequately evaluated because the training and evaluation data are not adequately separated.
As discussed in detail in Section 2.2, inappropriate data splitting can result in similarities between training and evaluation data (i.e., overtraining).
This often results in reported identification performance that exceeds the actual capabilities of the model.
We address and discuss this issue as in research question 1 (RQ1).

Second, we investigate the benefit of ROI extraction for pest identification in RQ2.
When dealing with fine-grained tasks like pest identification, it is very important to suppress model overfitting.
The pre-extraction of regions of interest (ROIs---leaf, fruit, etc.) has had some success in plant disease diagnostic tasks \cite{Saikawa2019} but has proven limited in its effectiveness due to the large environmental gap (domain gap) among fields \cite{Shibuya2021}.
Although a few studies have utilized pre-ROI detection in plant pest identification \cite{Li2019, Tassis2021}, to the best of authors' knowledge, no studies thus far have systematically validated the effectiveness of ROI identification.
If ROI identification can be adequately evaluated for its usefulness in plant pest discrimination, it will be helpful in the development of future integrated diagnostic techniques for identifying plant diseases and pests.
%

Third, we investigate the effectiveness of integrating similar pest types and introducing cross-crop training in the construction of pest discriminators.
Defining the composition of pest classes is also important in building pest identification models.
In the pest identification task, the difficulty of classifying different but visually similar pest species into the same taxon has been a significant obstacle to improving model performance.
However, from a practical standpoint, there is often little need to differentiate at the species level because the pest control methods employed (e.g., insecticides) do not differ among closely related species.
In other words, combining them into one category is considered an effective strategy.
Furthermore, when images of multiple crops damaged by the same pest species are available, combining the images of all crops into one massive class may allow the identification model to focus on the characteristics of the pest damage itself, independent of the crop.

In summary, this paper first explores the three RQs summarized below to develop a robust and practical plant pest identification model.

\begin{enumerate}
\item[RQ1:] How much does the adequate separation of training and test data affect performance?
\item[RQ2:] How effective is the pre-extraction of ROIs?
\item[RQ3:] Is pest/crop integration effective for practical systems?
\end{enumerate}
Based on the knowledge gained from working to answer the three RQs, we develop a practical plant pest identification model with high discriminative power and robustness using 334K highly growth-controlled real field images, consisting of 20 pests across four crops captured by agricultural experts in 27 fields from 2017 to 2021.

We assert that the primary audience for this paper is a wide range of researchers working in agriculture.
Therefore, we use common names for the pests analyzed, and corresponding scientific names are provided in the Appendix.

\section{Preparation: Strategy and three research questions}
In this section, we first review the methodology of machine learning (ML) techniques for automated plant pest identification, and we then introduce each RQ and related research that we identified in this study.
%

\subsection{Model selection for plant pest identification}
Most existing methods for diagnosing plant diseases and identifying pests from images utilize convolutional neural networks (CNNs) \cite{He2016, Tan2019} and object detection models \cite{Ren2017}.
Specifically, object detection models are capable of inferring both the location and the type of the target object, such as lesions, insect bodies, or signs of insect feeding.
CNN models that perform classification on an image-by-image basis are used in most studies because they have two advantages: a low preparation cost for training labels (i.e., only one label per image) and the ability to learn damage patterns without clear contours.
However, the small targets and subtle features used for identification in the plant pest dataset are at risk of being heavily obstructed by the noise and background \cite{Liu2021}.
In addition, CNNs are inherently susceptible to variations in distance from the object \cite{xu2014scale}.
Recently, object detection models have also been used in pest identification 
\cite{Liu2016, Fuentes2017, Selvaraj2019, Wang2020, Lin2020, Guo2021, Rong2022, Wang2021},
 as they are suitable for identifying objects with distinct contours, such as insect bodies, and they have the great advantage of being robust to variations in distance from the object.
These techniques are also used to estimate the extent of pest damage by counting the identified objects \cite{Li2019, Tetila2019, Guo2021, Rong2022}.

While object detection models are useful, they have limitations when it comes to identifying damage characterized by changes in the shape, color, or texture of the leaf or fruit. Extracting specific regions of damage is also a challenge for these models. 
Further, the cost of annotation can be immense, particularly when dealing with clusters of tiny insects.
Recognizing these issues, Suwa et al. \cite{Suwa2019} proposed a two-stage method for plant disease diagnosis. Their method first extracts regions of interest (ROIs), followed by disease diagnosis; this reduces both the risk of overfitting and the annotation costs associated with object detection models.
Drawing from this approach, we aim to build a two-stage pest identification model using CNNs.
Our method ensures that human annotation costs remain low, which is essential when dealing with large-scale data. 
%
The proposed framework is illustrated in Figure 3. The detailed explanation is given in Section 5.2.
%

\subsection{Impact of training and test data separation policy (RQ1)}
%
While numerous researchers in the field of plant pest identification have reported identification accuracies of 90\% or over on small training datasets \cite{Liu2016, Tetila2019, Tassis2021}, there are serious concerns regarding the generalizability of the models arising from a lack of rigorous separation of photographic contexts. 
In RQ1, we investigate the impact of data splitting strategies on measured model performance and discuss the partitioning criteria necessary to mitigate such issues.

Most previous studies have relied on cross-validation and random-splitting to divide a dataset into the training and test subsets.
Although convenient, these strategies may lead to situations in which the training and test sets share features beyond pest damages, thereby leading to overfitting on a specific set of conditions.
This concern has been confirmed by empirical studies in plant disease diagnosis \cite{Saikawa2019, Shibuya2021,Suwa2019, Mohanty2016, Ferentinos2018, Cap2020, Kanno2020, Guth2023}, as they report a significant reduction in the diagnostic ability of models when tested with images taken in environments different from the training environment.
Guth et al. \cite{Guth2023} reported that model generalizability assessments should be performed under diverse conditions using robust datasets.
%
There are a variety of factors that can cause overfitting of plant disease and pest classification models, such as image composition, brightness, and background, which can combine and potentially influence the model's judgment.
%
%
Figure 2 shows sample images of tomato whiteflies taken in five different fields (a) to (e).
%
It can be seen that, while images from the same field are not identical, they still possess similar features like the background and lighting conditions.
Meanwhile, images from different fields tend to have more diverse characteristics.
Some studies have attempted to counteract such effects; for example, Wu et al. \cite{Wu2023} applied domain adaptation techniques to models trained on images from a laboratory environment, and reported some success in improving the performance on images of plants in the field.
However, they also noted that their effectiveness is highly dependent on the test dataset.
Shibuya et al. \cite{Shibuya2021} reported a decline in diagnostic performance on images from fields unseen during training, despite efforts to maintain consistent data quality and characteristics, such as collecting images under similar composition and lighting conditions, and performing background removal.
The critical factors for this deterioration were not identified in the study.
%
%
Based on these findings, we hypothesize that separation of data sources is necessary in order for the test set to be a good representation of real-world images and analyze the relationship between data partitioning criteria and model performance through comparative experiments in RQ1.
In this study, we define “data sources” primarily as fields of image capture, as images from different agricultural fields are unlikely to share any characteristic apart from pest damages.
%

%
%
\begin{figure*}
    \centering
    \begin{minipage}{0.97\textwidth}
    \centering
    \includegraphics[width=\textwidth]{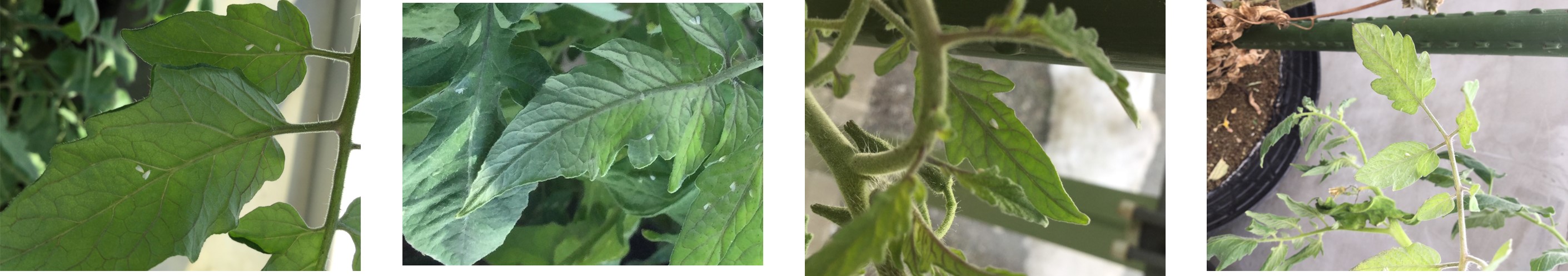}
    \caption*{(a) Images taken from farm field A}
    \end{minipage}
    \hfill
    \begin{minipage}{0.97\textwidth}
    \centering
    \includegraphics[width=\textwidth]{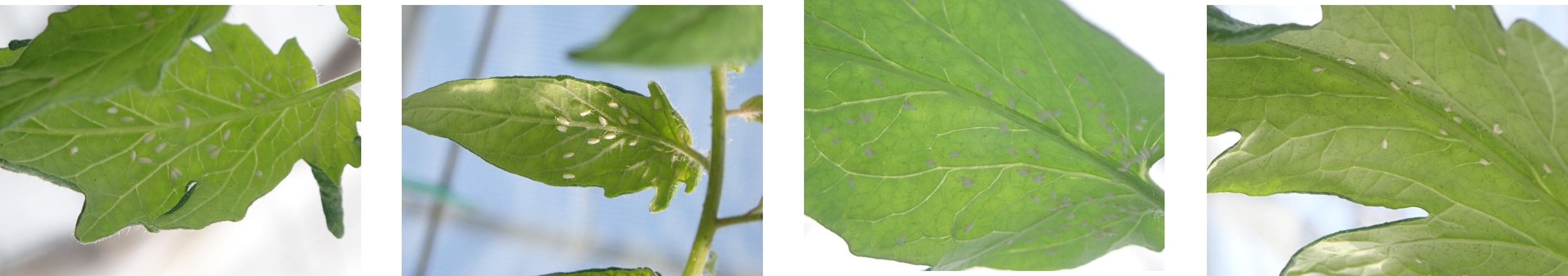}
    \caption*{(b) Images taken from farm field B}
    \end{minipage}
        \hfill
    \begin{minipage}{0.97\textwidth}
    \centering
    \includegraphics[width=\textwidth]{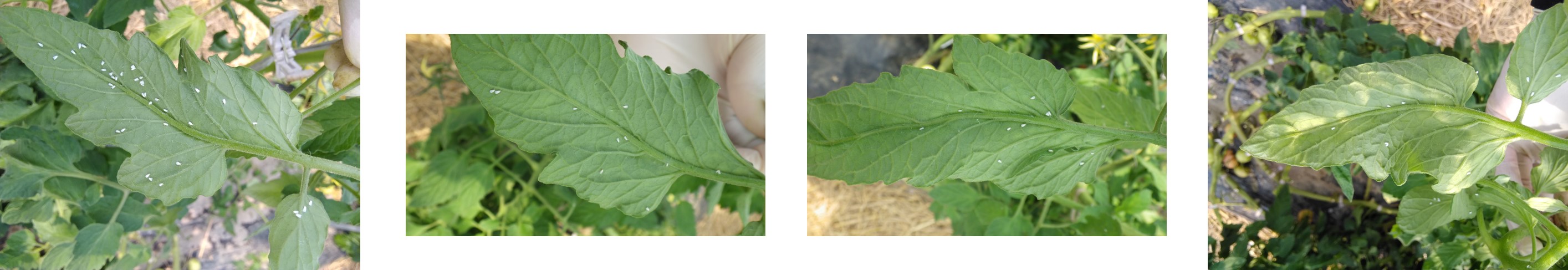}
    \caption*{(c) Images taken from farm field C}
    \end{minipage}
        \hfill
    \begin{minipage}{0.97\textwidth}
    \centering
    \includegraphics[width=\textwidth]{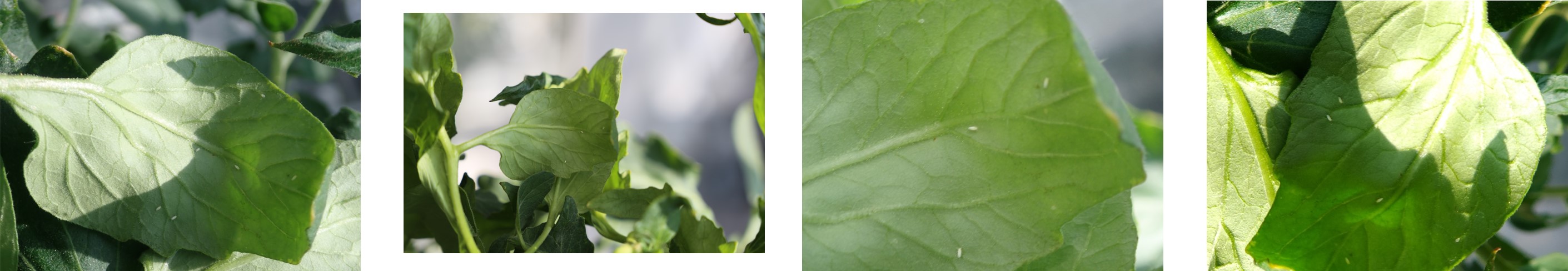}
    \caption*{(d) Images taken from farm field D}
    \end{minipage}
        \hfill
    \begin{minipage}{0.97\textwidth}
    \centering
    \includegraphics[width=\textwidth]{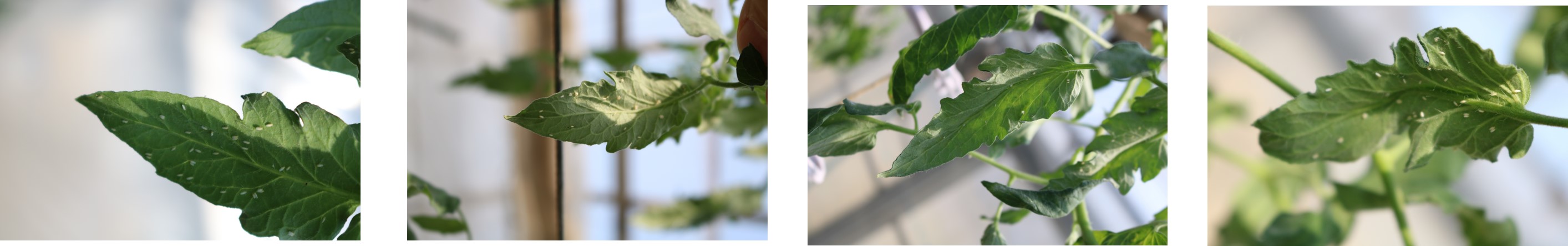}
    \caption*{(e) Images taken from farm field E}
    \end{minipage}
    \caption{Example of images taken from different farm fields (whiteflies on tomato leaf). \\ 
    In the 'same farm' scenario of RQ1, test images were randomly sampled from the images from farm fields (a)--(e), and the remainder were used for training. In the 'different farm' scenario, images from farms (a)--(c) constituted the training set, while those from (d)-(e) were used for evaluation.}
    \label{fig:fig2}
\end{figure*}


\begin{figure*}
    \centering
    \includegraphics[width=0.97\textwidth]{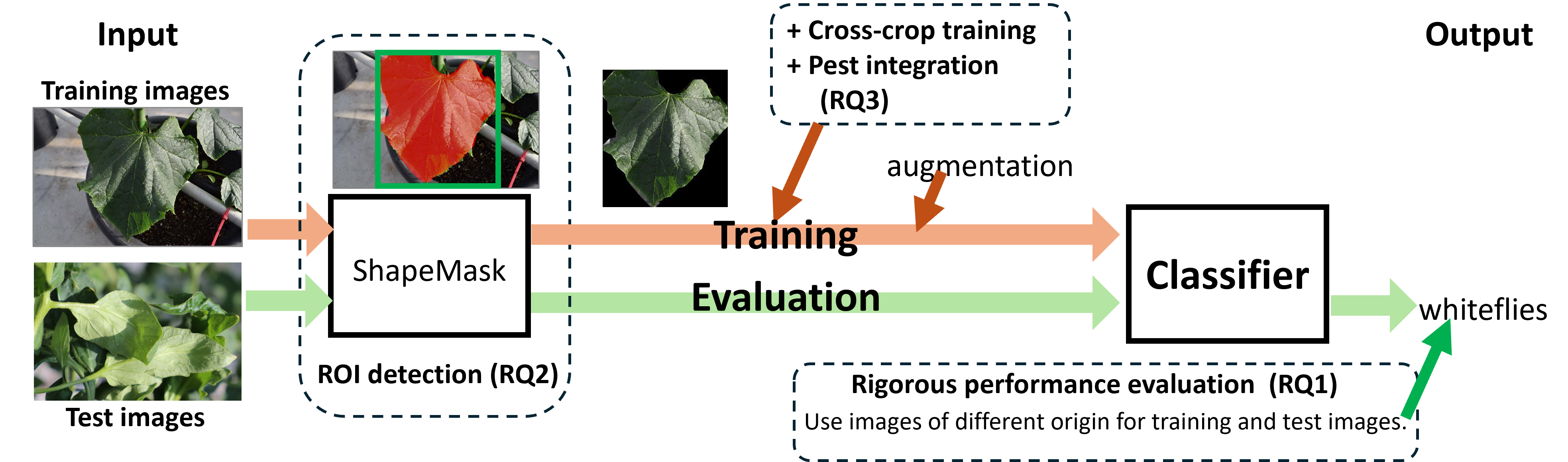}
    \caption{Schematics of our two-stage plant pest identifier}
    \label{fig:fig3}
\end{figure*}

\begin{figure}
\begin{subfigmatrix}{4}
  \subfigure[Kanzawa spider mite]{\includegraphics[width=0.24\textwidth]{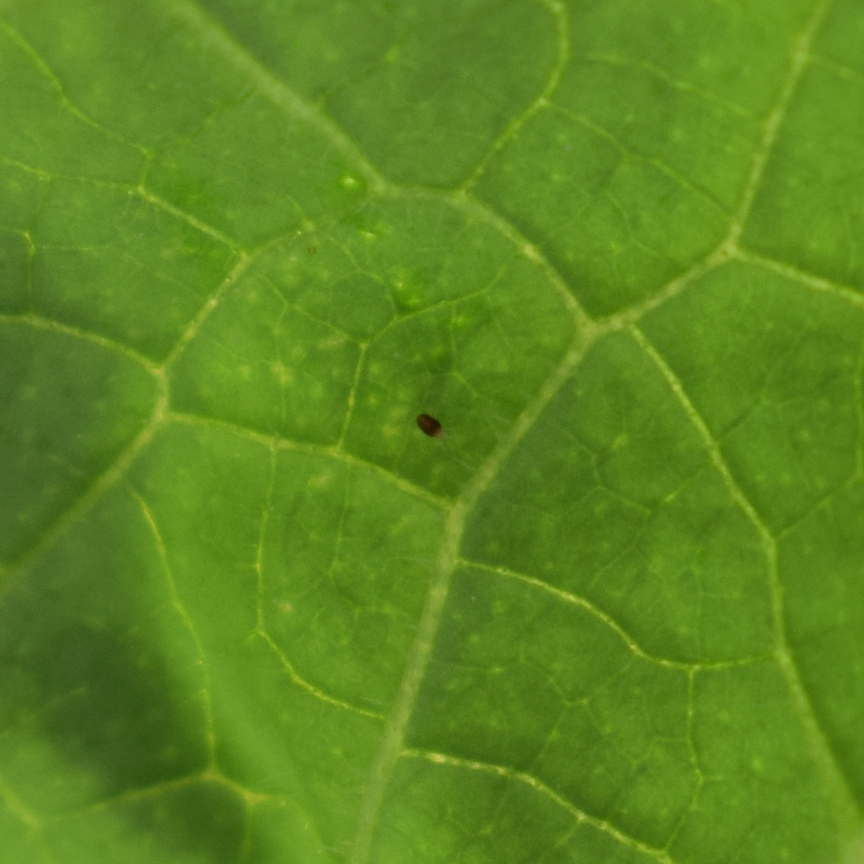}}
  \subfigure[Twospotted spider mite]{\includegraphics[width=0.24\textwidth]{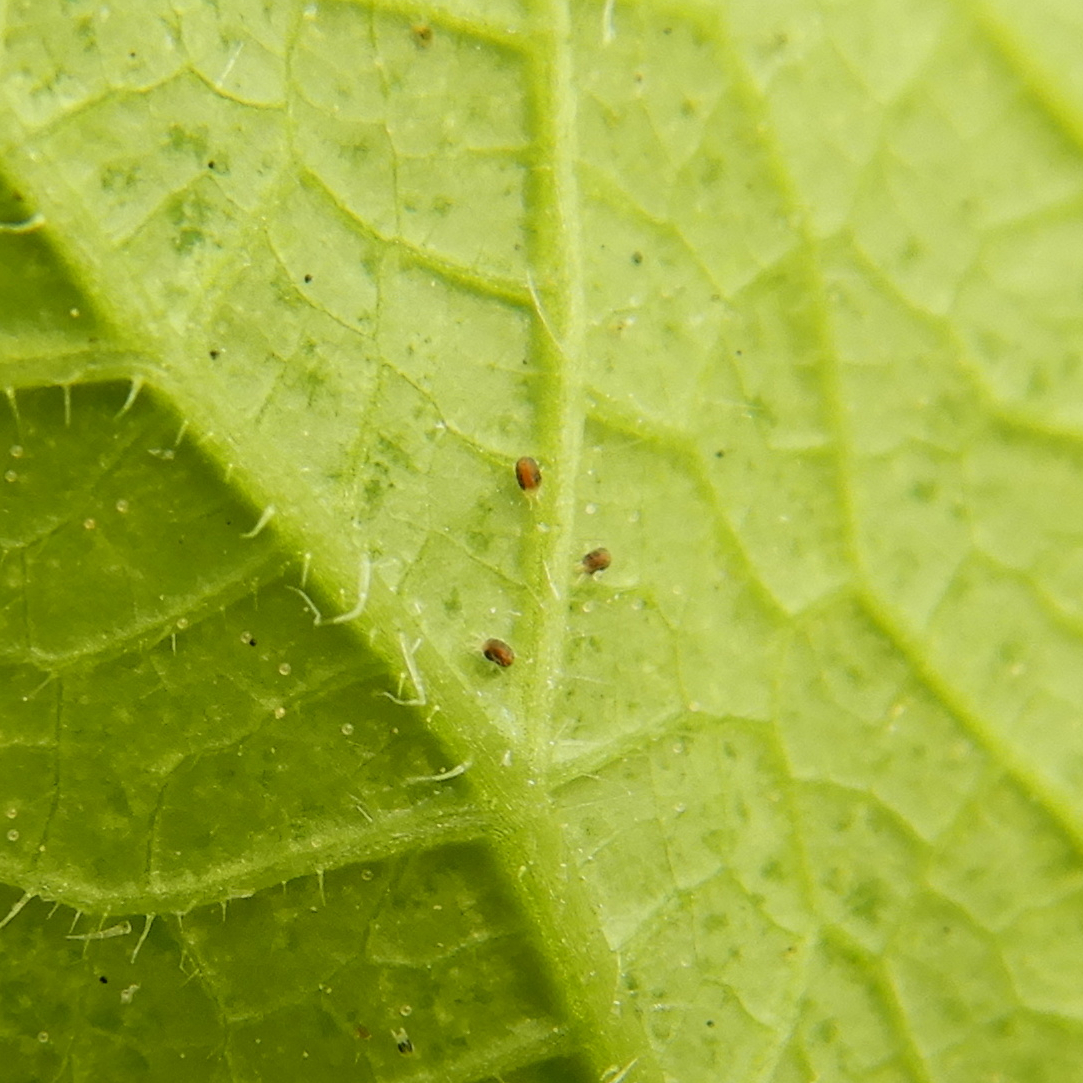}}
\subfigure[Cotton aphid]{\includegraphics[width=0.24\textwidth]{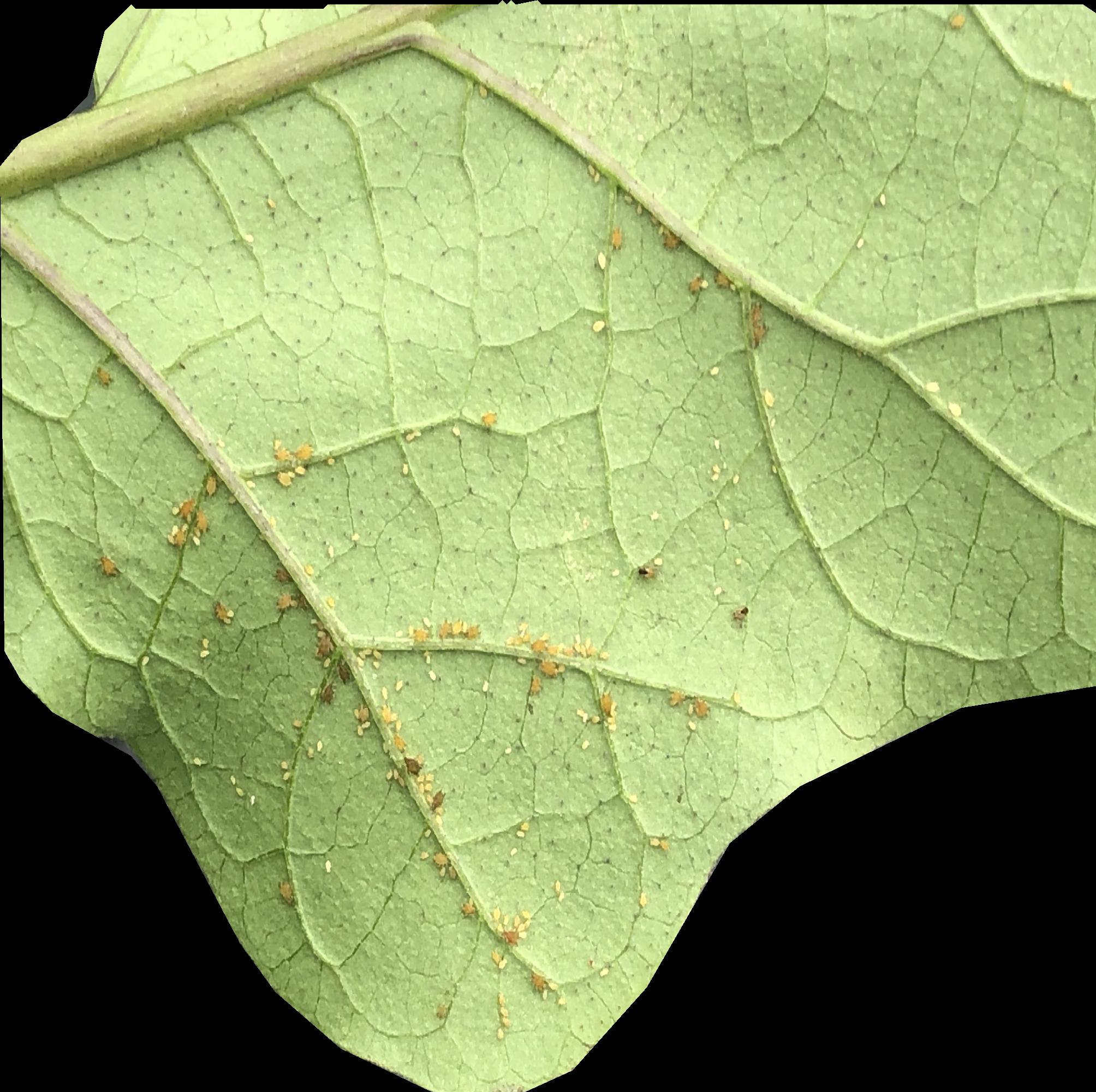}}
\subfigure[Greep peach aphid]{\includegraphics[width=0.24\textwidth]{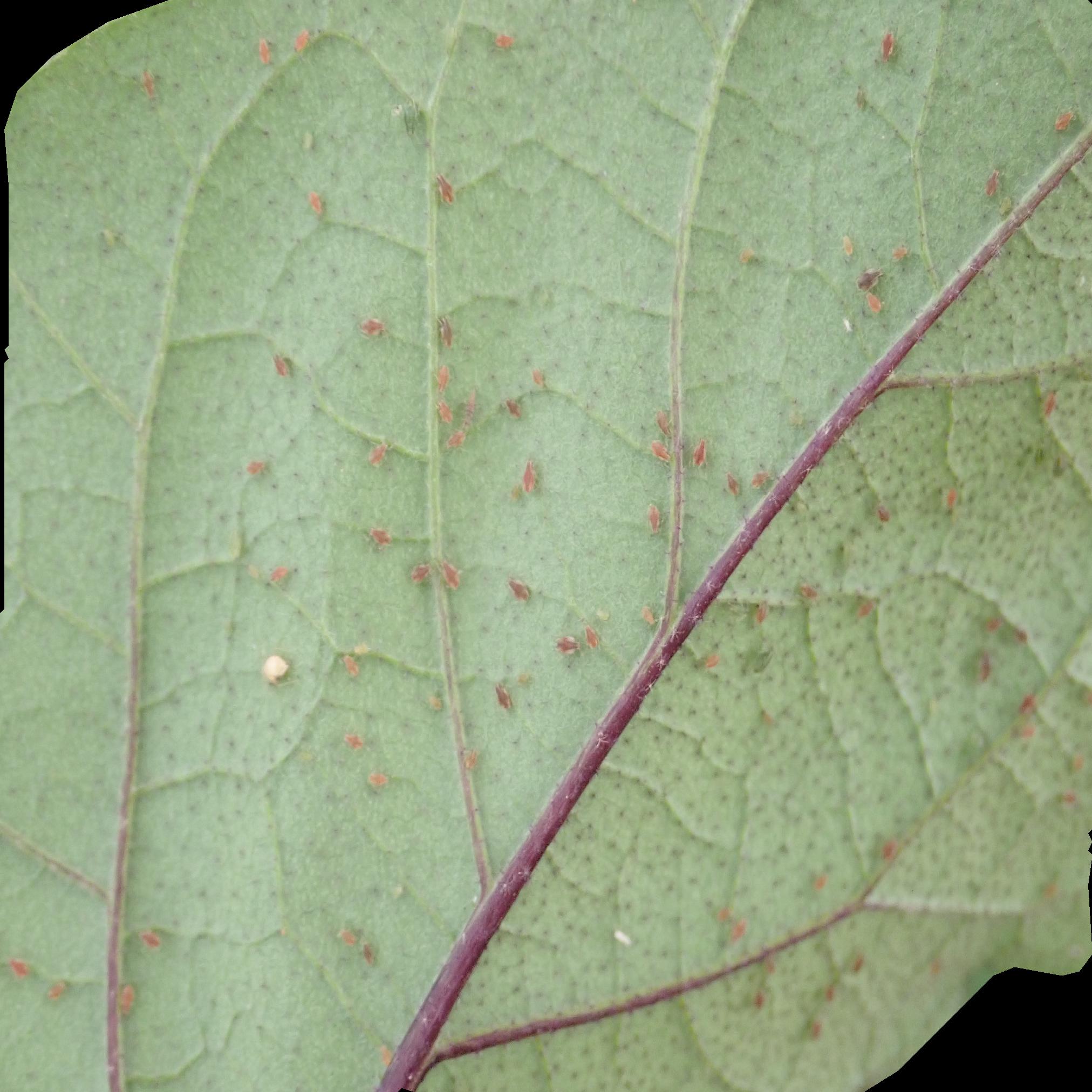}}
\end{subfigmatrix}
\caption{Example of two spider mites on cucumber (a)(b) and two aphids on eggplant (c)(d) which are difficult to discriminate. \\  {\footnotesize These images are cropped and enlarged for visualization. The impact of integrating each of these spider mite and aphid classes on identifiability was evaluated using RQ3.}}
\label{fig:fig4}
\end{figure}

\subsection{Impact of ROI detection (RQ2)}
In computer vision tasks, even in the era of deep learning, ROI preselection remains a common method to achieve a greater performance in certain applications.
In the field of plant disease diagnosis, it has been asserted that the background can affect diagnostic performance \cite{Liu2021, Saikawa2019, Shibuya2021}, and directing the discriminator's attention to the object to be identified is considered effective in improving performance.
Saikawa et al. \cite{Saikawa2019} built a highly accurate leaf region detector using pix2pix \cite{Isola2017} and confirmed an improvement in disease diagnosis accuracy with the elimination of background information.
However, a subsequent validation based on higher-resolution images and more sophisticated CNN models also suggested that in the disease diagnosis task, ROI pre-extraction alone is insufficient to deal with the aforementioned overlearning problem, as the ROI itself---such as leaves---preserves domain-specific information \cite{Shibuya2021}.
Unlike disease indicators, plant pest cues often involve identifying smaller, localized objects, and the intra-species diversity tends to be less than that in diseases.
Given these characteristics, ROI detection could prove effective in plant pest identification.
However, this concept has not been explored well in quantitative studies.
To answer this RQ, we investigate the effectiveness of ROI detection.

\subsection{Impact of class definition (RQ3)}
Most automated diagnostic studies adopted the method of class division as a given condition, and insufficient attention has been paid to which taxonomic level (species, genus, family, etc.) is practically appropriate to form a single class or to whether multiple crops infested by the same pest species should be treated together in the same class.
Here, we discuss two factors that affect the performance of identification models: 1) the combination of closely related pests and 2) the use of multiple crop-damaging pest images together for training.
First, certain species have minute differences in appearance, as depicted in Figure 4, but images taken with common smartphones or other devices indicate few distinct differences.
Conversely, closely related pest species are often treated with similar control methods; thus, there is little merit in distinguishing them as separate pest species.
Second, it is common for a pest species to infest more than one crop.
Consequently, it would be a substantial contribution to practical modeling if a model's performance could be improved by training on images of damage caused by the same pest but on different plant species.
The first topic, the integration of similar pest classes, has been addressed in previous studies
\cite{Kong2022, Liu2021},
but the effects of doing so have not been evaluated or discussed.
Meanwhile, the second, cross-crop training, has not been systematically addressed in any study.
Therefore, in this RQ, we investigate the impact of classification definitions on identification performance of leaf pests from numerous images.

\section{Material and settings}
This section describes the overall dataset and pest identification framework utilized in this study.

\subsection{Data collection}
In this study, images of four plant portions (front and back of the leaf, fruit, and flower) of each of four crops (tomato, strawberry, cucumber, and eggplant) taken across 27 farms in total, including agricultural experiment stations in 24 prefectures in Japan, were used.
These consist of 78 crop portion–pest combinations in total, including 20 pest species and one healthy category for each crop.
Various digital cameras and smartphones were used to collect images of insect bodies and damage marks (e.g., feeding damage marks) manually, and each image was tagged with metadata, including crop, location, pest species, plant portion, and date and time of capture.
Pest species were identified by experts at each institution and used as a gold standard label, whereas images featuring multiple types of pest damage were not included in this study and were excluded in advance.
The image data used in our study will be made available sequentially on the website 
(\url{https://github.com/ai-pest/pest_damage_image_db/}).

To develop a practical model and conduct a meaningful analysis, two conditions were set for image collection: 
(i) Images should be taken in such a manner that the body of the insect or some other trace that is evidence of insect damage is roughly centered in the image.
Macro photography was not used at this time (as presented in Figure 1).
(ii) Images should be taken during daylight hours to avoid insufficient light.

The images used in this experiment have the following three characteristics:
(1) They are rich in diversity.
Photos taken by different photographers in different farm fields are included in the same category (i.e., class), thereby allowing for a wide variety of backgrounds and subjects for each class. 
(2) Many of the damaged areas and insects in the images we processed are smaller in size than existing plant pest datasets; moreover,  because a single leaf or fruit was photographed to fit within the image, the damaged area appears smaller than in macrophotographed images of insect bodies and feeding scars. 
(3) The number of images of the crop and their respective portions depends on the type of pest species (Tables 1, 2, 3, and 5).
While images of healthy plants can be obtained throughout the year, images of pest-infested plants can only be captured during the periods when the specific pests are active.
Certain species are particularly scarce because their infestation levels are strongly influenced by the natural environment. 

Note that data collection and discriminator development were conducted in parallel over the five years of this research project (2017--2022). 
Therefore, the number of images differs between the experiments corresponding to each RQ presented in Section 4 and the main experiment presented in Section 5.
The specific number of classes and images used is described in each experiment.

\subsection{Elimination of inappropriate images}
Some of the captured images were mislabeled, out of focus, or taken sequentially by the cameras burst mode, all of which can negatively affect model development.
Therefore, a combination of automatic detection and visual inspection was adopted to exclude images unsuitable for training.
First, to exclude sequential images that were highly similar to each other, images taken at intervals of less than one second were excluded by referring to the image timestamps.
Next, similar images were extracted and removed using the features extracted by an image classification model.
Specifically, ResNet101 \cite{He2016}---pre-trained on ImageNet provided by FastAI \cite{Howard2020}---was fine-tuned with the collected images, and a 512-dimensional representation was obtained from the hidden FC layer immediately before the output layer.
Thereafter, the low-dimensional representations of the images were compiled into a database, and similar images were extracted and removed using the nearest neighbor search system, FAISS \cite{Johnson2021}. 
The automatically filtered images were then visually inspected for further refinement, and blurred, close-up, and other low-quality images were excluded from the dataset.

\subsection{ROI detection}
The ROI detector used in this paper was ShapeMask, which generates high-resolution masks and is reported to have high performance \cite{Kuo2019}.
In addition, we used ResNet101 (image resolution: 256$\times$256 pixels) pre-trained on ImageNet as the backbone of ShapeMask and fine-tuned it with polygonal annotations.
This enabled the robust extraction of target locations, such as leaves and fruits, from the original image at runtime.

To train ShapeMask, the ROI of each instance was annotated as polygons (see Figure 3).
Because creating high-resolution polygon-type annotations is extremely labor-intensive, they were generated using a semi-automated method based on self-training.
First, polygons were manually created for several hundred of the collected images (e.g., 100--400 images) for use in the initial stage of training the ShapeMask model, which in turn generated polygons for other images, totaling several times the number of samples in the training data (approximately 1,000 images).
Then, the generated polygons were visually inspected, corrected, and used to retrain the model along with the existing training data.
By repeating this cycle, an annotated dataset of images with polygons surrounding ROIs was progressively assembled.
When this ROI detection result was subsequently used in the discriminator, a rectangular area image bounded by the ShapeMask was extracted and scaled up or down to a 1:1 aspect ratio, and the background was replaced with black, as depicted in Figure 3.
This resizing process prevents the resolution of the image inputted into the discriminator from being reduced when the detected leaves and fruits are small in the original image, and allow the microscopic insect bodies and feeding damage marks to be inputted into the discriminator at a relatively high resolution.

\subsection{Classifiers and evaluation criteria}
In this study, EfficientNet \cite{Tan2019}, a CNN model that achieves superior discriminative power with fewer parameters, was used as a pest discriminator in all experiments.
The details of the model, including the image resolution, are discussed in the respective sections. The performance of the pest discrimination in each RQ and the main results were measured using the average micro accuracy, and the F1 score (harmonic mean of precision and recall) for each category, and the macro F1 score.

\section{RQ initiatives and their results}
In this section, we describe the experimental details and results corresponding to the three RQs.
As mentioned above, our research project spanned five years---from 2017 to 2022.
Within this time frame, as data were collected, three RQs were investigated individually.
Therefore, different datasets were used for different purposes to answer each RQ.
Finally, for all RQs, we focused on analysis on images of leaves (front and back), as they allow for more extensive data collection and a wider variety of pests compared to images of other plant portions.

\subsection{RQ1: Training and test data separation policy}
\subsubsection{RQ1: Conditions}
To investigate the effect of the policy of separating training and testing data on the identification performance, two scenarios were prepared.
The “same farm" scenario represents the method used in most previous studies, where the data set is constructed by merging images from single or multiple farms, which are then randomly separated for training and evaluation.
In other words, images from the same farm field can be included in both the training and evaluation data.
%
Meanwhile, in the “different farms" scenario, the dataset was partitioned so that images from the same farm field are represented in only one of the subsets.
This splitting method enables the measurement of more intrinsic performances.
The left portion of Table 1 presents the image data used in the experiments, which consisted of leaf images from 13 different pest groups---including healthy ones---from four crops.
In both scenarios, the number of test images was the same.
%
For the tomato whiteflies shown earlier in Figure 2, in the same farm scenario, training images were randomly sampled from the images (a)--(e), and the rest were used for evaluation.
In the different farm scenario, images (a)--(c) were used for training, and images (d) and (e) were used for evaluation.
The discriminative model used was EfficientNet-B5, and the resolution of the input images was set to 256 pixels square.
Training was performed for 160 epochs and the data augmentation methods used were horizontal flipping, vertical flipping, and rotation (random, every 90 degrees).
Each of these three pre-processing types was applied independently and with equal probability (50\% each with and without inversion, and 25\% per 90-degree rotation).
Note that no ROI pre-detection was performed for this RQ.

\begin{table}[t]
\begin{center}
\caption{Data used in RQ1 (data segregation policy) and corresponding results}
\small
\begin{tabular}{ l r r c r r} \hline
\multirow{2}{*}{Pest class $\dag$} & \multicolumn{2}{c}{\# images} & & \multicolumn{2}{c}{Identification in F1 (\%)}\\ \cline{2-3}\cline{5-6}
 & training & test & & same farm & \bf{different farm} \\ \hline
C\_whitefly& 556 & 395 & &97.3 & 25.0\\
C\_thrips& 876 & 67 & &96.4 & 41.5\\ 
C\_healthy& 859 & 99 & &91.3 & 6.5\\
E\_broad mite& 947 & 4 & &53.3 & 3.8 \\
E\_thrips& 636 & 316 & &96.8 & 41.0\\
E\_leafminer& 652 & 300 & &96.8 & 55.7\\
E\_healthy& 704 & 244 & &96.7 & 43.4\\
S\_spider mite& 502 & 107 & &95.0 & 57.1\\
S\_cotton aphid& 776 & 165 & &95.7 & 46.3 \\
S\_tobacco cutworm& 649 & 271 & &96.2 & 44.9 \\
S\_healthy& 868 & 122 & &88.0 & 26.7 \\
T\_whitefly& 255 & 35 & &86.2 & 12.4 \\
T\_healthy& 881 & 65 & &85.1 & 38.7\\ \hline
Total & 9,161 & 2,190 & & & \\ 
Macro F1 (\%) &  &  & &90.4 & 34.1\\ 
Accuracy (\%) & & & &95.2 & 36.5 \\ \hline 
\end{tabular}
\end{center}
{\small $\dag$: leaf images. C: cucumber, E: eggplant, S: strawberry, and T: tomato; twospotted spider mite is abbreviated to spider mite.\\}
\end{table}

\subsubsection{RQ1: Results}
The identification results are also presented in Table 1. 
The mean accuracy and macro F1 for the model were 95.2\% and 90.4\%, respectively, in the same farm scenario, whereas they were significantly lower in the different farm scenario (36.5\% and 34.1\%).
Moreover, the F1 scores per class deteriorated in the different farm scenario in all 13 classes.
The model's poor performance in the different farm scenario reveals its inability to handle farm-to-farm variations, an issue not evident in the metrics from the same farm scenario. 
Although the same farm scenario results align with trends observed in previous studies, these findings may not provide an accurately representation of the model's real-world performance, thereby revealing a common oversight of practical generalizability in existing research.
%
%
Looking again at Figure 2, the images used for evaluation in the different farm scenario (d, e) have larger brightness differences within the leaf area than the images used for training (a)--(c), and the contrast of the insect body is relatively low.
Although the training images contained various images that captured the insect body well, the diverseness of the images was insufficient for the difficulty of the problem.
As a result, in the test images (d, e), there were many misidentifications to other pest damage learned from images with large differences in brightness or missed insect bodies.

Further, it is important to note that the F1 score for the “eggplant\_broad mite" was particularly low because the number of test images was very small, and the precision was greatly reduced due to false positives from other classes (36.4\% in the same farm, but only 2.0\% in the different farm scenario). 
Conversely, the recall was comparable to other classes (100\% in the same farm scenario, 75.0\% in the different farm scenario).
This is a common problem when class imbalances are significant.

\begin{table}
\begin{center}
\caption{Data used in RQ2 (ROI detection) and corresponding results}
\small
\begin{tabular}{ l r r c r r} \hline
\multirow{2}{*}{Pest class} & \multicolumn{2}{c}{\# images} & &\multicolumn{2}{c}{Identification in F1 (\%)}\\ \cline{2-3}\cline{5-6}
 & training & test & & baseline & \bf{MASKED} ${}^\dag$ \\ \hline
Broad mite & 2,357 & 83 & &36.4 & 42.4\\
Spider mite & 9,977 & 1,558 & &79.2 & 80.4\\
Whitefly & 6,092 & 188 & &70.7 & 70.1\\
Aphid & 11,954 & 1,917 & &88.6 & 89.2\\ 
Thrips & 9,976 & 2,021 & &79.4 & 81.6\\
Hadda beetle & 3,377 & 1,029 & &60.6 & 68.4 \\
Leafminers & 1,485 & 223 & &40.6 & 50.7 \\
Cotton bollworm & 2,107 & 30 & &23.7 & 32.5 \\
Tobacco cutworm & 8,256 & 1,588 & &83.6 & 84.7\\
Healthy & 52,092 & 1,271 & &68.4 & 75.4\\ \hline
Total & 107,673 & 9,908 & & &\\
Macro F1 (\%) &  &  & &63.1 & 67.5\\
Accuracy (\%) & & & &76.0 & 79.3 \\ \hline
\end{tabular}
\end{center}
{\small $\dag$: Images were subjected to preliminary ROI detection using the method described in Section 3.3.\\}
\end{table}

\subsection{RQ2: ROI detection}
\subsubsection{RQ2: Conditions}
To evaluate the effect of ROI pre-detection, we compared the discriminative ability of two scenarios (“masked" and “baseline") with and without ROI pre-detection.
%
The left portion of Table 2 presents the breakdown of the dataset.
A total of 107,673 images were used for training, comprising 10 classes of healthy and pest-damaged leaves of 4 crops (tomato, cucumber, eggplant, and strawberry).
A total of 9,908 eggplant images from 10 classes were used for evaluation.
%
In this experiment, some pest species were integrated (see RQ3).
In the evaluation of the “masked" scenario, ROI detection was also performed using the evaluation images.
In all subsequent experiments, including those related to RQ2, the evaluation images were, in principle, captured in a field different from that in the training images (i.e., similar to that in the different farm scenario of RQ1).
However, for classes with a small number of farm fields (aphids, thrips, spider mites, leaf miners, and whiteflies), it was difficult to collect a sufficient number of images; therefore, we divided images from the same field into training and evaluation subsets, provided the image capture date was different.
This was done to ensure an adequate number of images for training and evaluation, while minimizing the effects of overlearning.
The identification model used was EfficientNet-B1, and the input resolution was 1024 pixels square.
The training duration was set to 30 epochs, and image scaling and Mixup \cite{Zhang2018} were applied as data augmentation methods, in addition to horizontal and vertical flipping and rotation, as described in RQ1.
For image scaling, the central region of the input image (80\% of the height and width of the input image, respectively) is cropped and enlarged to 1024 pixel squares by linear interpolation.
For image reduction, a 102-pixel border is created on all four sides of the input image, the image is reduced to a 1024-pixel square by linear interpolation and the borders are blackened.
The maximum alpha value for Mixup is set to 0.2.
In this RQ, 12,187 images from the training data were utilized to train the ShapeMask model for ROI detection. Visual inspection confirmed that ROIs were correctly detected.
In addition to quantitative analysis, Grad-CAM \cite{Selvaraju2017} is used in this experiment to visualize the model's ROI in the identification and to investigate how EfficientNet's ROI changes with and without background removal.

\subsubsection{RQ2: Results}
A summary of the identification results is presented in the right portion of Table 2.
The model in the “masked” scenario (i.e., with ROI pre-detection) demonstrated a 3.3-point higher accuracy and 4.4-point higher macro F1 score than the baseline, at 79.3\% and 67.5\%, respectively.
Meanwhile, the F1 scores per class were higher in the masked scenario in 9 of the 10 classes, except for whiteflies.
Figure 5 presents a comparison of the attention focus, indicated by Grad-CAM between the baseline and masked models for an image of an eggplant leaf with thrips.
The baseline model without ROI detection misidentified the subject of the image as a broad mite, but the masked model achieved accurate identification.
For the broad mite and cotton bollworm classes, the F1 score was less than 50\%, although there was improved performance for ROI detection.
As with RQ1, this is largely due to a significant decrease in precision due to false identification results from other classes, as the number of images evaluated is smaller than those in other classes.
Specifically, the ROI extraction yielded the following results: broad mite recall: 83.1\% $\rightarrow$ 84.3\%, precision: 23.3\% $\rightarrow$ 28.3\% and cotton bollworm recall 23.3\% $\rightarrow$ 43.3\%, precision: 24.1\% $\rightarrow$ 26.0\%.
Although the identification performance of these two classes was poor, ROI detection improved the recall of cotton bollworm by 20\%, which was particularly difficult for the model to identify.

\begin{figure*}[ht]
\begin{subfigmatrix}{5}
 \subfigure[Original input]{\includegraphics[width=0.19\textwidth]{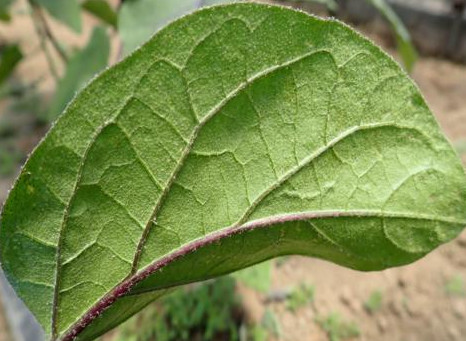}}
 \subfigure[Grad-CAM by baseline model (misidentified as broad mite)]{\includegraphics[width=0.19\textwidth]{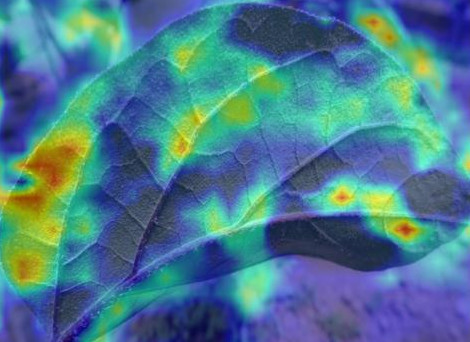}}
 \subfigure[Masked input]{\includegraphics[width=0.19\textwidth]{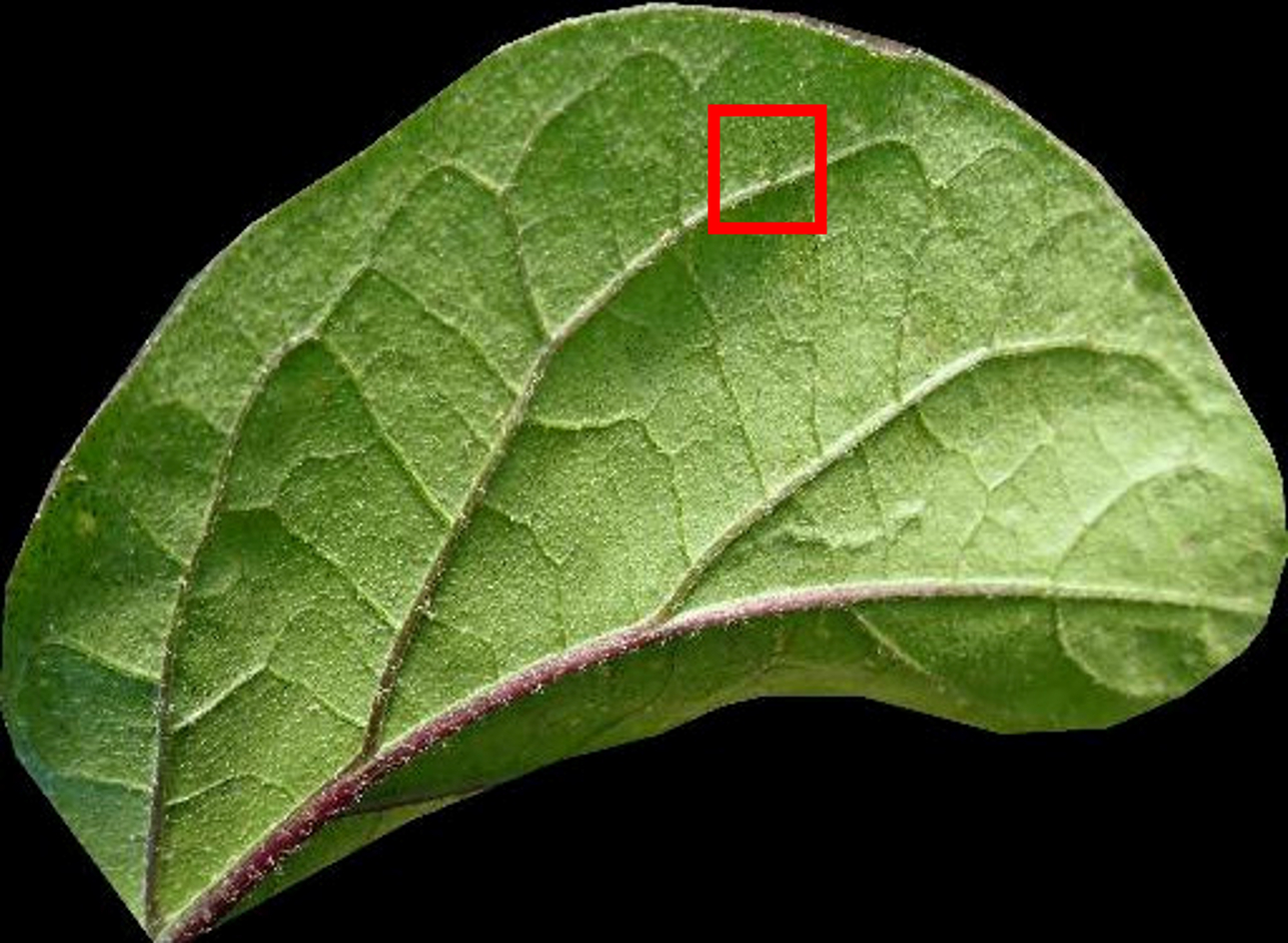}}
 \subfigure[Grad-CAM by masked model (correctly identified as thrips)]{\includegraphics[width=0.19\textwidth]{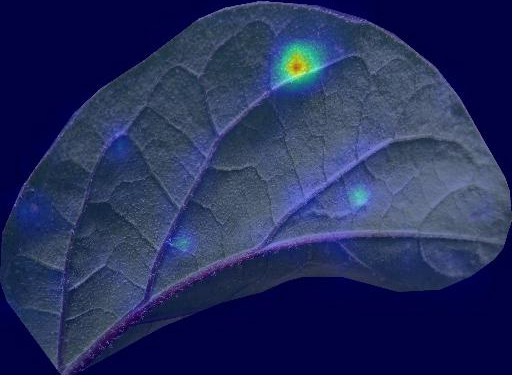}}
 \subfigure[Enlarged insect body (c)]{\includegraphics[width=0.13\textwidth]{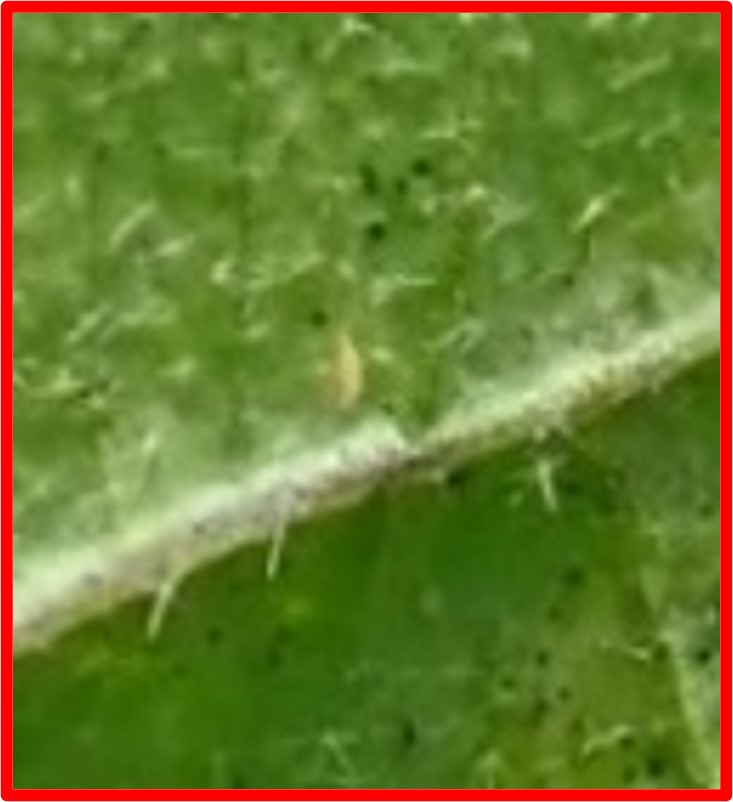}}
\end{subfigmatrix}
\caption{Differences of model attention on pest identification with and without ROI detection (eggplant leaf with thrips).}
\label{fig:fig5}
\end{figure*}

\subsection{RQ3: Class definition}
\subsubsection{RQ3: Conditions}
The effects of (A) “integrating closely related species" (pest integration) and (B) “adding images of other crops" (cross-crop training) on identification performance were investigated.
In scenario (A), pests are integrated as one class if the same pesticides can be applied and if they are taxonomically related.
In scenario (B), if available, instances of other crop damage by the same pest species are added as supplement training data for each class in the dataset. 
Table 3 summarizes the number of images used in this experiment.
The baseline dataset consists of seven classes of cucumber leaf images (Table 3a) and ten classes of eggplant leaf images (Table 3b).
For cucumber (Table 3a), the identification performances were evaluated and compared in five scenarios: baseline, (A) pest integration, (B1) cross-crop training (healthy only), (B2) cross-crop training (full), and (C) both pest integration and cross-crop training (combined).
For eggplant (Table 3b), only the effect of (A) pest integration was evaluated.

In the pest integration scenario (A), two closely related spider mite species, Kanzawa spider mite (KSM) and twospotted spider mite (TSM), were combined into a single class named "spider mites" on cucumber and eggplant. Similarly, two aphid species---cotton aphid and green peach aphid---were integrated into the class "Aphids" on eggplant (see Figure 4).

In the cross-crop scenario (B) on cucumber, images of three crops (tomato, strawberry, eggplant) are added to some classes of the training images.
This scenario consists of two sub-scenarios, full cross-crop and healthy cross-crop training.
In the (B2) full cross-crop sub-scenario, approximately 49,000 images from other crops are added to three pest classes (thrips, spider mites, and healthy) that had below-average accuracy in the baseline model evaluation.
In the (B1) healthy cross-crop sub-scenario, only the “healthy” class receives additional data from other crops. 
The primary aim of this sub-scenario is to evaluate the effectiveness of adding images of healthy cases from a variety of crops, which are more readily available than those of pest-infested crops.
In the combined scenario (C), both (A) and (B2) are applied.
%
Note that the purpose of including images of different crops in this cross-crop training here is to obtain common features from a limited amount of training data.
We believe that, in principle, it is desirable to build a discriminator for each crop.

In all these scenarios, ROI detection was performed using the method described in Section 3.3, where the training and evaluation images were basically divided by the field of origin; however, two classes with a significant bias in the number of images per field (broad mite and KSM) were divided on the condition that the images were taken on different dates.
EfficientNet-B1 was used as the classification model, with the resolution of the input images set to 1,024 $\times$ 1,024 pixels; the models were trained for 30 epochs, and horizontal and vertical flipping and image rotation were applied as data augmentation processes, as in RQ1.

\begin{table*}[ht]
\caption{Data used in RQ3 (class definition)}
\begin{center}
%
\subtable[Cucumber images used in RQ3;  baseline, (A) pest-integration, (B) cross-crop training, and (C) combined scenarios]{
\footnotesize
\begin{tabular}{ l r r r r r c r r } \hline
\multirow{3}{*}{Pest class} & \multicolumn{5}{c}{\# training images} && \multicolumn{2}{c}{\# test images }\\ \cline{2-6} \cline{8-9}
 & Baseline & (A) Integ. &\multicolumn{2}{c}{(B) Cross-crop training} & (C) Combined && Baseline & (A) Integ.\\
  & & &  (B1: healthy only) & (B2: full) & (A+B2) &&  (B) cross-crop & (C) combined \\ \hline
Broad mite & 1,031 & 1,031 & 1,031 & 1,031 & 1,031 && 823& 823\\
KSM & 1,452 & \multirow{2}{*}{$\rbrace$ 4,625${}^\ast$}& 1,452& 1,452& \multirow{2}{*}{$\rbrace$ 9,306${}^\ast$}&& 1,141 & \multirow{2}{*}{$\rbrace$ 1,371${}^\ast$}\\
TSM${}^\dag$ & 3,173& & 3,173 & 7,854 & && 230 & \\
Whitefly & 5,353 & 5,353 & 5,353 & 5,353 & 5,353 && 1,435 & 1,435 \\
Cotton aphid & 2,056 & 2,056 & 2,056 & 2,056 & 2,056 && 222 & 222\\
Thrips${}^\dag$ & 3,332 & 3,332 & 3,332 & 12,686 & 12,686 && 319 & 319\\
Healthy${}^\dag$ & 17,010 & 12,010 & 51,719 & 51,719 & 51,719 && 1,358 & 1,358\\ \hline
Total & 33,407 & 33,407 & 68,116 & 82,151 & 82,151 && 5,528 & 5,528\\ \hline
\end{tabular}
} 
\\

\subtable[Eggplant images used in RQ3;  baseline, and (A) pest-integration scenarios]{
\footnotesize
\begin{tabular}{ l c r r c r r } \hline
\multirow{2}{*}{Pest class} & \ \ \  & \multicolumn{2}{c}{\# training images} & \ \   & \multicolumn{2}{c}{\# test images }\\ \cline{3-4} \cline{6-7}
 && Baseline & (A) Integ. &&  Baseline & (A) Integ.\\ \hline
Broad mite && 1,300 & 1,300 && 37 & 37 \\ 
KSM && 1,900 & \multirow{2}{*}{$\rbrace$ 3,174${}^\ast$}&& 379 & \multirow{2}{*}{$\rbrace$ 492${}^\ast$}\\
TSM && 1,274& && 113 &   \\
Cotton aphid && 2,458 & \multirow{2}{*}{$\rbrace$ 6,725${}^{\ast\ast}$}&& 208 & \multirow{2}{*}{$\rbrace$ 519${}^{\ast\ast}$}\\
Green peach aphid && 4,267& && 311 &   \\
Thrips     && 4,629 & 4,629 && 691 & 691 \\ 
Leafminer && 912 & 912 && 765 & 765 \\
Tobacco cutworm && 4,542 & 4,542 && 1,461 & 1,461 \\
Hadda beetle && 3,766 & 3,766 && 1,015 & 1,015 \\ 
Healthy && 11,317 & 11,317 && 1,007 & 1,007 \\ \hline
Total && 36,365 & 36,365 && 5,987 & 5,987\\ \hline
\end{tabular}
} 
\end{center}
{\footnotesize
KSM: Kanzawa spider mite, TSM: twospotted spider mite\\
  ${}^\ast$ KSM and TSM are integrated into “spider mite" in the pest-integration scenario (A).\\
  ${}^{\ast\ast}$ Cotton aphid and Green peach aphid are integrated into “aphid" in the pest-integration scenario (A).\\
  ${}^\dag$ Images of additional crops were included in the learning in the cross-crop training scenario (B).\\
  }
\end{table*}

\begin{table*}[t]
\begin{center}
\caption{Results of RQ3}
\subtable[Effect of (A) pest integration and (B) cross-crop training on cucumber]{
\footnotesize
\begin{tabular}{ l r r r r r} \hline
\multirow{3}{*}{Pest class} & \multicolumn{5}{c}{Identification performance in F1 (\%)}\\
 & Baseline & (A) Pest integration &\multicolumn{2}{c}{(B) Cross-crop training} & (C) Combined \\ 
 & & & (B1: healthy only) & (B2: full) & (A+B2)\\ \hline
Broad mite & 69.9 & 70.8 & 74.8 & 73.3 & 71.8 \\
KSM & 78.7 &\multirow{2}{*}{$\rbrace$ 90.6}& 79.7 & 81.4 & \multirow{2}{*}{$\rbrace$ 90.1}\\
TSM & 37.2 && 37.0 & 45.3 & \\
Whitefly & 99.0 & 99.2& 99.0 & 98.9 & 99.1 \\
Cotton aphid & 88.6 & 89.3 & 89.6 & 90.1 & 91.7 \\
Thrips & 71.2 & 70.9 & 69.8 & 68.5 & 71.6 \\
Healthy & 71.9 & 73.9 & 73.6 & 72.7 & 73.9 \\ \hline
Macro F1 (\%) & 73.8 & 82.4 & 74.8 & 75.8 & \bf{83.0} \\
Accuracy (\%) & 78.6 & 84.6 & 79.7 & 80.0 & \bf{84.7} \\ \hline
\end{tabular}
} 
%
\subtable[Effect of (A) pest integration on eggplant]{
\footnotesize
\begin{tabular}{ l c r r} \hline
\multirow{2}{*}{pest class} && \multicolumn{2}{c}{Identification performance in F1 (\%)}\\
 && Baseline & (A) Pest integration \\ \hline 
Broad mite && 46.0 & 48.4 \\ 
KSM && 79.5 & \multirow{2}{*}{$\rbrace$ 85.4}\\
TSM && 30.5&  \\
Cotton aphid && 59.1 & \multirow{2}{*}{$\rbrace$ 82.1}\\
Green peach aphid && 49.0 &   \\
Thrips     && 70.3 & 74.3 \\ 
Leafminer && 87.4 & 86.9 \\
Tobacco cutworm && 86.2 & 85.7 \\
Hadda beetle && 70.3 & 69.5 \\ 
Healthy && 90.3 & 91.0 \\ \hline
Macro F1 (\%) && 66.9 & \bf{77.9} \\
Accuracy (\%) && 78.2 & \bf{82.7} \\ \hline
\end{tabular}
} 
\end{center}
{\footnotesize KSM: Kanzawa spider mite, TSM: Twospotted spider mite\\}
\end{table*}

\subsubsection{RQ3: Results}
The evaluation results are presented in Table 4.
%
In the pest integration scenario (A), the “spider mite" class with integration in both crops and the “aphid" class with integration in eggplant achieved significantly better practical performance than without integration.
It is also noteworthy that the performances of the other classes that were not integrated did not deteriorate significantly.
Consequently, the models in the integrated scenarios showed remarkable improvement in macro F1 scores (8.6\% for cucumber and 11.0 \% for eggplant) from respective baseline models.

Of two (B) cross-crop scenarios with cucumber images, the full cross-crop (B2) scenario improved in terms of accuracy and macro F1 score by 1.4 and 2.0 points, respectively.
Further, an improvement in per-class F1 scores was observed across all classification categories except thrips and whiteflies, thereby confirming the effectiveness of the method.
The healthy cross-crop model (B1), where only healthy images of other crops are added to the training images, outperformed the baseline by 1.1 points in accuracy and 1.0 points in macro F1 score, and there was a modest increase in per-class F1 scores in the healthy and broad mite classes. 
In the combined scenario (C; A+B2), the accuracy and macro F1 score improved by 6.1 and 9.2 points, respectively, over the baseline. 
Although the degree of performance improvement of the full combined scenario was minimal compared to that of the full cross-crop scenario (0.1 points in accuracy, 0.6 points in macro F1 score), the effectiveness of the strategy was confirmed, primarily in poorly performing categories (thrips and broad mites).

\subsection{Summary of findings answering for the RQs}
Here, we briefly summarize the findings regarding the three RQs. 
A more detailed discussion is provided in the Discussion section.
The above experiments reveal that in automatic plant pest identification,
(1) the performance evaluation should assess the generalizability of the model.
To this end, the images in the test set should be sufficiently considered to have diverse characteristics that differ from those in the training set, such as separate source farms or images taken on different days.
%
(2) Pre-extraction of ROIs---such as leaves and fruits---helps to direct the attention of the model towards the object that needs to be identified, thereby improving identification accuracy.
Meanwhile, (3a) the integration of pest categories with similar visual characteristics and pesticide susceptibility is a promising option for model improvement, as the technique also mitigates the adverse effects of the scarcity of evaluation data on measured model performance.
Finally, (3b) the performance of the discriminator can be improved by including images of different crops with pest damage to the dataset, provided that the damage patterns are comparable in terms of visual characteristics.
%


\section{Develop of an accurate identification model based on RQ answers}
In this section, we describe the development process and results of a robust, high-accuracy plant pest identification model.
This model, which is the main result of this paper, was developed utilizing the findings from the three RQs and the entirety of the obtained large-scale data.

\subsection{Training and evaluation images}
A summary of the training and evaluation images is presented in the left portion of Table 5.
In this experiment, we collected 334,314 images across 27 fields, including agricultural experiment stations, in 24 Japanese prefectures.
The images depict both pest-damaged and healthy portions of four different crops: tomato, strawberry, cucumber, and eggplant. Specifically, we captured images of leaves (both front and back), fruits, and flowers.
Then, we filtered the dataset as described in Section 3.2 to exclude images inappropriate for training and evaluation, thereby reducing the number of samples to 279,810.
Lastly, ROI extraction was applied to obtain 331,194 images for analysis, which were used in the experiment.
Note that multiple ROIs may exist within an image, and each ROI was converted into a separate image in this process; therefore, the number of samples increased after ROI extraction.
Based on the results for RQ1, the training and evaluation data were separated in order to ensure that the fields from which they were taken would be strictly different.
As with RQ2, ROI extraction was performed on both training and evaluation images.
In addition, the integration of closely related species and multiple cultures (RQ3) was also included in this experiment.
Each class in the dataset follows a naming convention in the format “Species (portion\_crop),” where the 'portion' refers to the plant portion infested (like fruit or leaf), and 'crop' refers to the specific crop affected.
If a portion or a crop is not specified, it indicates that the images encompass multiple portions or crops for that particular pest species.

In this study, there are a total of 20 pest species as shown in the "common name" column of Table A in the Appendix and the total number of combinations of $\{$pest, portion, crop$\}$ was 78.
Similar pest species, as well as portions and crops of the same pest species, were integrated to develop a total of 18 $\{$pest, portion, crop$\}$ classes (IDs 1-18).
This was combined with 7 classes of pest-free crop and portion combinations (IDs 19-25) to develop a total of 25 classes of datasets.
%
To increase the evaluation's rigor further, classes with insufficient images---which could not be divided into training and test subsets---were only used for training and excluded from the evaluation.
Therefore, the training data for this experiment includes 25 classes, while the evaluation data only contains 21 of these classes.

\subsection{Classification framework}
%
%
Here we explain our proposed framework for plant pest identification, shown in Figure 3.
Based on the findings from RQ2, we employed a two-stage identification framework that combines a ROI detection model and a CNN-based image classifier to achieve practical plant pest identification with a high accuracy and robustness.
A total of 461,517 polygon annotations for 298,128 images were used to train the ShapeMask for ROI detection according to the method described in Section 3.3.
Meanwhile, the model used for identification was EfficientNet-B6, with a modified resolution of 1,024 $\times$ 1,024 pixels and which was pre-trained on ImageNet and JFT-300M \cite{Sun2017} and retrained with the collected images.
During training, the batch size was set to 64, the training period was set to 30 epochs, and the stochastic gradient descent (SGD) was used as the optimizer.
%
%
Each ROI-detected image is subjected to the cross-crop training and the pest integration, which were confirmed to be effective in RQ3, and used to train the discriminator along with data augmentation.

The data augmentation methods utilized were horizontal and vertical flipping and rotation, as described in RQ1, as well as image scaling and reduction, as described in RQ2.
%
%
Examples of augmented images are shown in Figure 6.
%
\begin{figure*}
    \centering
    \includegraphics[width=0.97\textwidth]{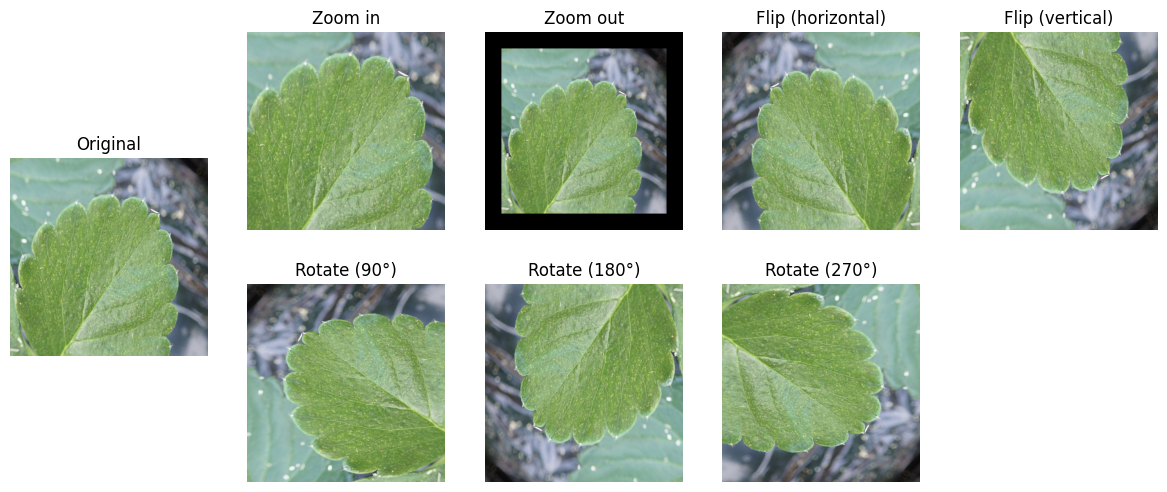}
    \caption{Example of data augmentation}
    \label{fig:fig6}
\end{figure*}
%
%
%

\begin{table}
\begin{center}
\caption{Data statistics in main experiments and main results}
\small
\begin{tabular}{c l r r c r r r} \hline
\multirow{2}{*}{ID} & \multirow{2}{*}{Pest class ${}^\dag$} & \multicolumn{2}{c}{\# of images} & &\multicolumn{3}{c}{Identification performance (\%)}\\
 & & \# train & \# test &  & Recall &  Precision & F1\\ \hline
(1) & Broad mite (fruit\_strawberry) & 54 &  &  &  &  & \\
(2) & Broad mite (fruit\_eggplant) & 1,327 & 11 & & 100.0 & 73.3 & 84.6 \\
(3) & Broad mite (leaf) & 3,316 & 189 & & 74.1 & 87.5 & 80.2 \\
(4) & Spider mites (leaf) & 17,040 & 823 & & 80.2 & 92.3 & 85.8 \\
(5) & Tomato russet mite (leaf\_tomato) & 3,529 & 184 & & 65.8 & 90.3 & 76.1 \\
(6) & Whitefly & 18,279 & 684 & & 92.3 & 93.8 & 93.0 \\
(7) & Aphid & 33,443 & 1,319 & & 94.5 & 89.8 & 92.1 \\
(8) & Mealybug (fruit\_eggplant) & 51 &  &  &  &  &\\
(9) & Mealybug (leaf\_eggplant) & 174 &  &  &  &  &\\
(10) & Thrips (fruit\_strawberry) & 4,682 & 155 & & 87.1 & 100.0 & 93.1 \\
(11) & Thrips (fruit\_tomato) & 3,171 & 1,278 & & 75.1 & 98.8 & 85.3 \\
(12) & Thrips (flower\_strawberry) & 5,072 & 249 & & 96.0 & 98.4 & 97.2 \\
(13) & Thrips (leaf) & 18,601 & 867 & & 95.3 & 92.7 & 94.0 \\
(14) & Hadda beetle (leaf\_eggplant) & 6,127 & 99 & & 96.0 & 80.5 & 87.6 \\
(15) & Leafminer (leaf) & 3,319 & 324 & & 98.5 & 96.7 & 97.6 \\
(16) & Cotton bollworm (fruit\_tomato) & 1,189 & 126 & & 92.1 & 96.7 & 94.3 \\
(17) & Cotton bollworm (fruit\_eggplant) & 1,179 & 76 & & 97.4 & 94.9 & 96.1 \\
(18) & Tobacco cutworm (leaf) & 21,754 & 1,531 & & 87.6 & 97.7 & 92.4 \\
(19) & Healthy (fruit\_strawberry) & 9,138 & 27 & & 100.0 & 60.0 & 75.0 \\
(20) & Healthy (fruit\_cucumber) & 2,745 &  &  &  &  &\\
(21) & Healthy (fruit\_tomato) & 5,354 & 700 & & 98.6 & 68.9 & 81.1 \\
(22) & Healthy (fruit\_eggplant) & 3,556 & 154 & & 99.4 & 100.0 & 99.7 \\
(23) & Healthy (flower\_strawberry) & 7,172 & 15 & & 60.0 & 60.0 & 60.0 \\
(24) & Healthy (flower\_cucumber) & 2,223 & 117 & & 100.0 & 100.0 & 100.0 \\
(25) & Healthy (leaf) & 146,476 & 3,295 & & 97.6 & 90.6 & 94.0 \\ \hline
Total & & 318,971 & 12,223 & & & \\ 
Macro F1 (\%)& & & & & & & 88.5 \\
Accuracy (\%)& & & & & & & 91.0\\ \hline
\end{tabular}
\end{center}
{\small ${}^\dag$ pest classes are named in the format “{pest species} ({portions}\_{plant name})," and omitted cases have been integrated to reflect the results pertaining to RQ3 (78 classes $\rightarrow$ 25 classes).
For the rigorous identification performance evaluation, all test data images were captured in a different field from the one from which the training data were obtained (RQ1). 
Further, the model was trained on the integrated 25 classes, but four classes for which data were from different farms were unavailable (classes 1, 8, 9, and 20) were excluded from the evaluation.\\
All images are ROI-detected (RQ2).}
\end{table}

\subsection{Main results}
A summary of the evaluation results of this experiment is presented in the right portion of Table 5, and the confusion matrix normalized to the ground truth is depicted in Figure 7.
The accuracy and macro F1 score were 91.0\% and 88.5\%, respectively, and F1 scores per class were above 80\% for 18 of the 21 species, except for  “Tomato russet mite (leaf\_tomato)" (5), “Healthy (fruit\_strawberry)" (19), and “Healthy (flower\_strawberry)" (23)  of which 12 were above 90\%.

Unlike many of the previous studies, where training and evaluation images are collected from the same farm(s), this evaluation result reveals the practical performance of the model by testing it on images from unseen fields.
Therefore, we are confident that this model will perform well in real use cases, as our model is supported by a large dataset and effective strategies.
The average identification time was 476 ms/case with an Intel Xeon W-2123 and an NVIDIA Titan V GPU.
Our framework was able to identify plant pests accurately in most cases, even though clues for pest identification (e.g., insect bodies and feeding scars) are often minute.

Figure 8 depicts typical examples of correctly identified pest damage, along with their respective Grad-CAM attention maps.
By comparing the original images (the top row) and the images after ROI extraction (the middle row), we observe a substantial enhancement in the resolution of important features, such as insect bodies and damaged regions. 
This pre-processing pipeline assists the classifier by providing it with high-resolution images, which is particularly impactful when these features are small.
The attention maps reveal that the discriminator correctly identified the insect bodies or scars as unique features of the predicted classes.

Figure 9 presents examples of successful and failed identification cases in low discriminative performance categories, (3), (5), and (23), each corresponding to the arrow in the confusion matrix 
 in Figure 7.
In Figure 9(b), as indicated by the arrow (i) in the confusion matrix, the image of a healthy strawberry flower (23) was incorrectly identified as thrips (12). 
Similarly, in Figures 9 (d) and (f), corresponding to arrows (ii) and (iii) in the confusion matrix, broad mites (3) and tomato russet mites (5) were incorrectly identified as healthy leaves (25). 
These examples indicate that the model struggled to identify pests with particularly small bodies or faint damage signs, thereby failing to distinguish between healthy and damaged samples.
%

\begin{figure}[t]
    \centering
    \includegraphics[width=0.80\textwidth]{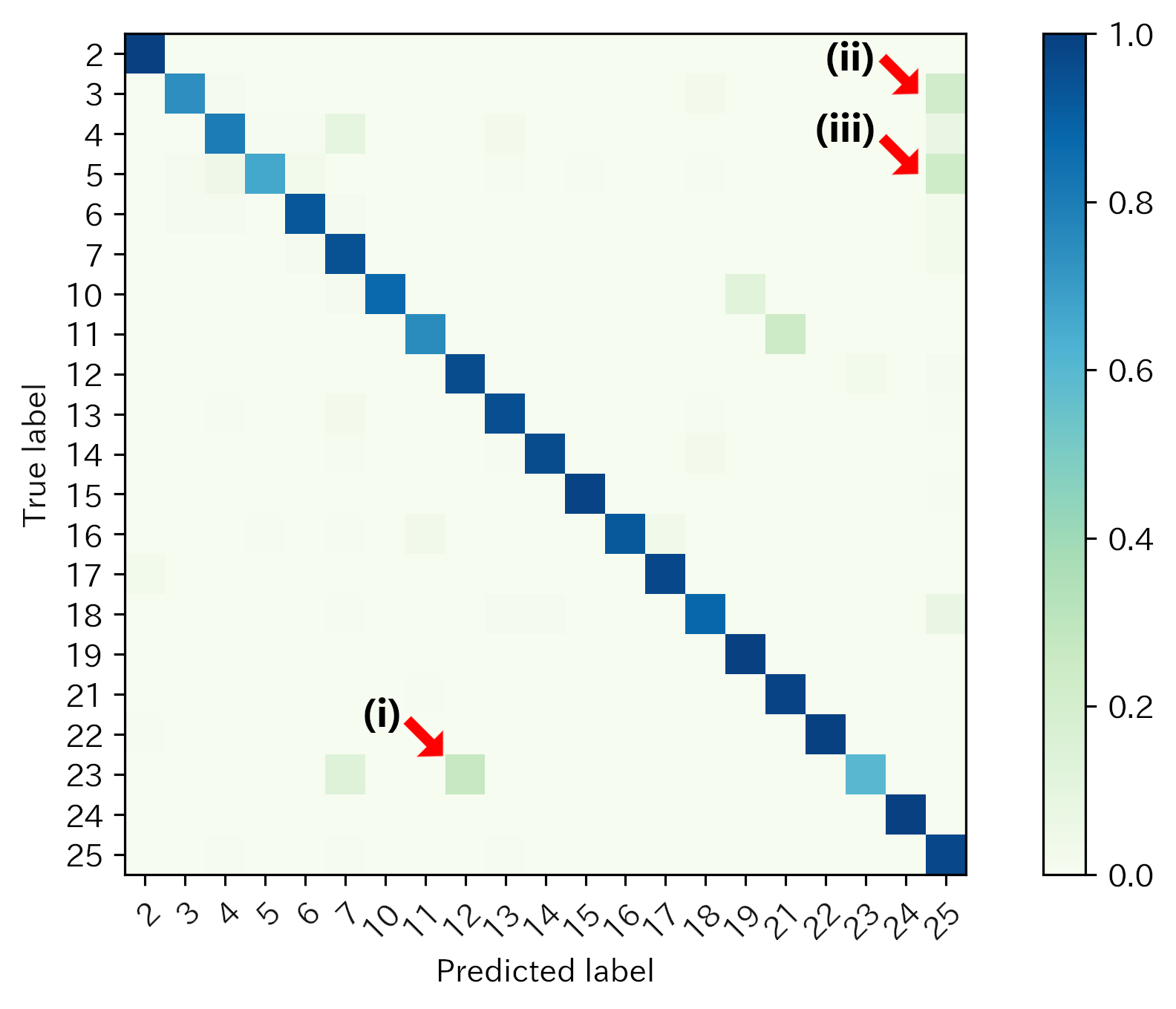}
    \caption{Confusion matrix of pest identification using the proposed framework normalized with recall. Specific examples of misidentified results for (i)--(iii) are depicted in Figures 9 (b), (d), and (f), respectively. Four classes of pests were excluded due to the lack of evaluation data.}
    \label{fig:fig7}
\end{figure}

\setcounter{subfigure}{-6} 
\begin{figure}
\begin{subfigmatrix}{3}
 \subfigure{\includegraphics[]{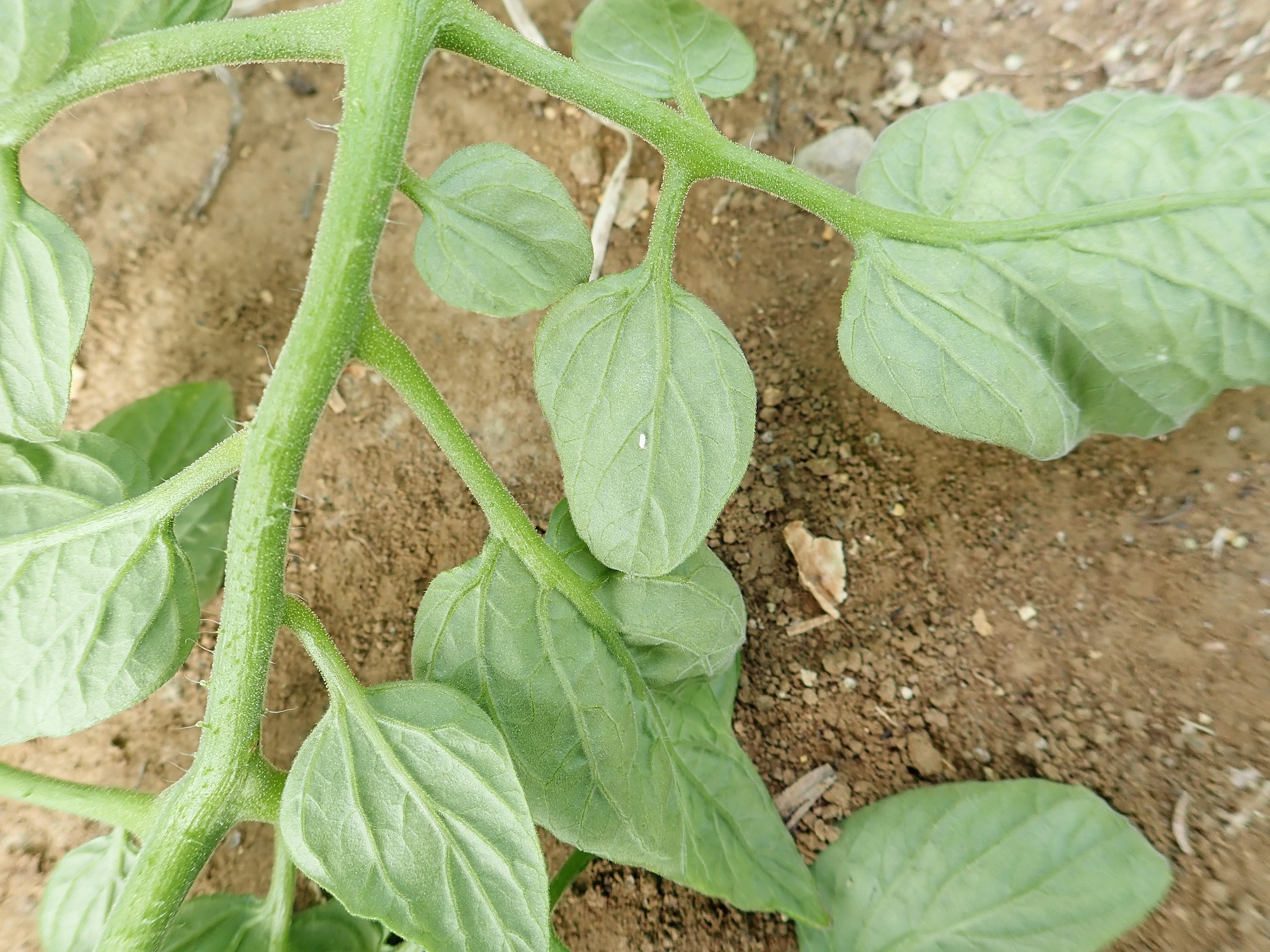}}
 \subfigure{\includegraphics[]{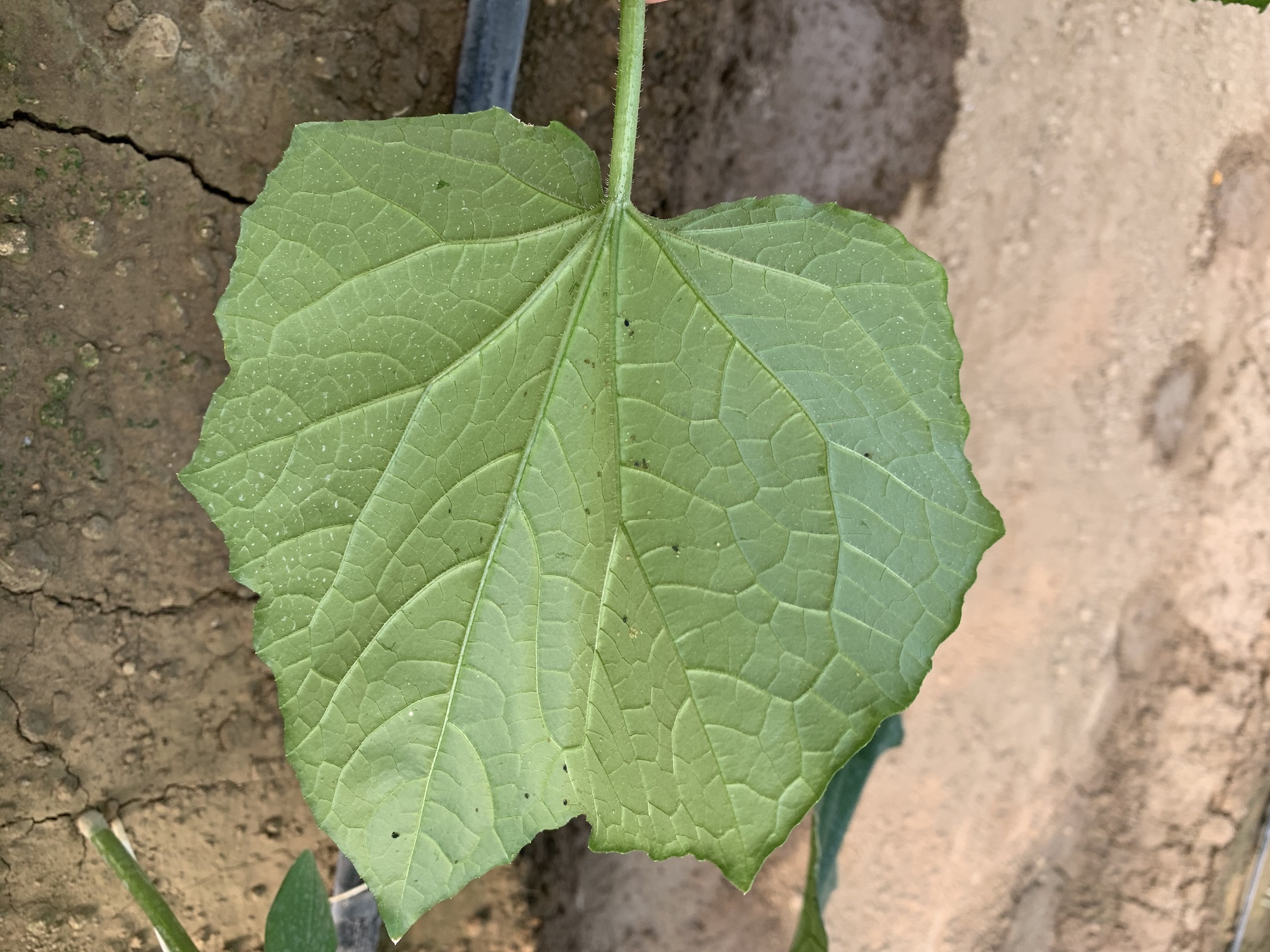}}
 \subfigure{\includegraphics[]{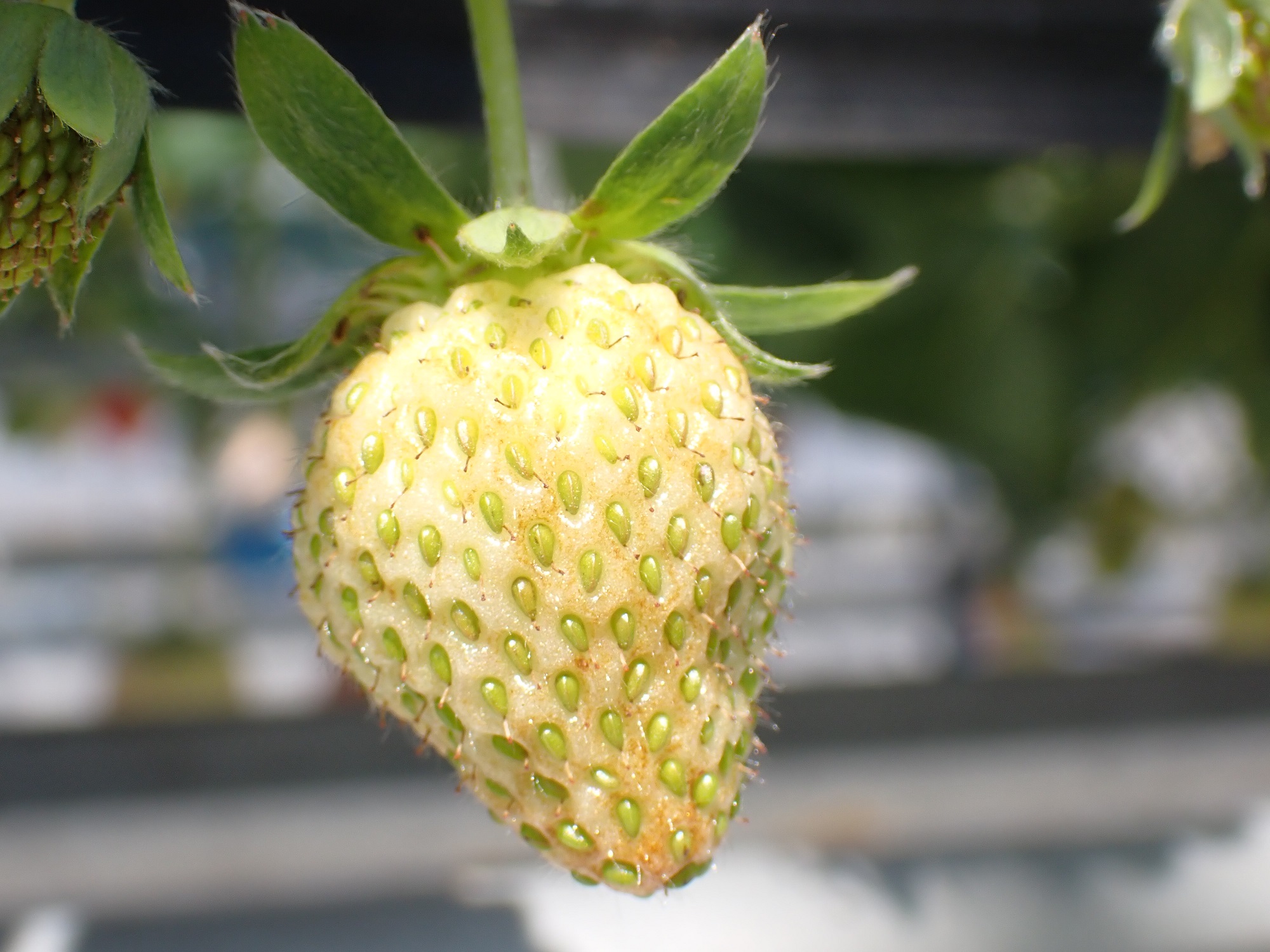}}
 \subfigure{\includegraphics[]{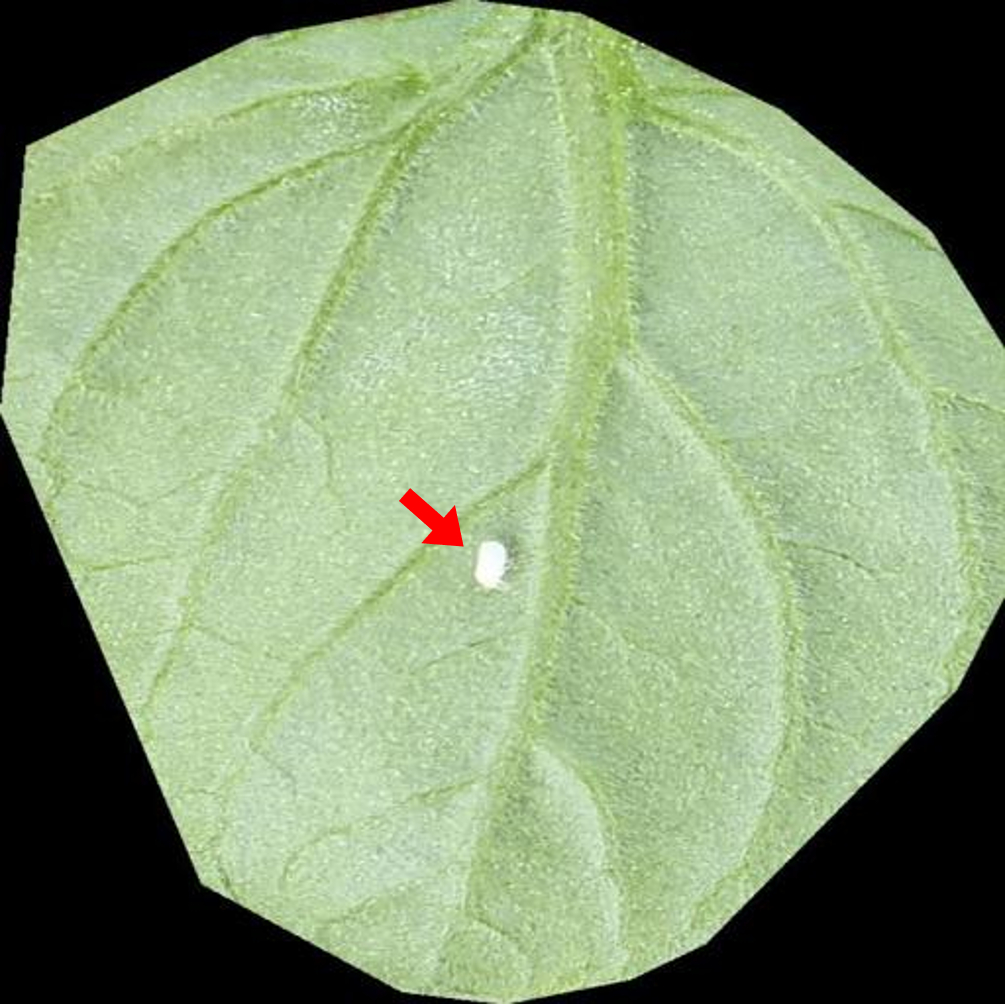}}
 \subfigure{\includegraphics[]{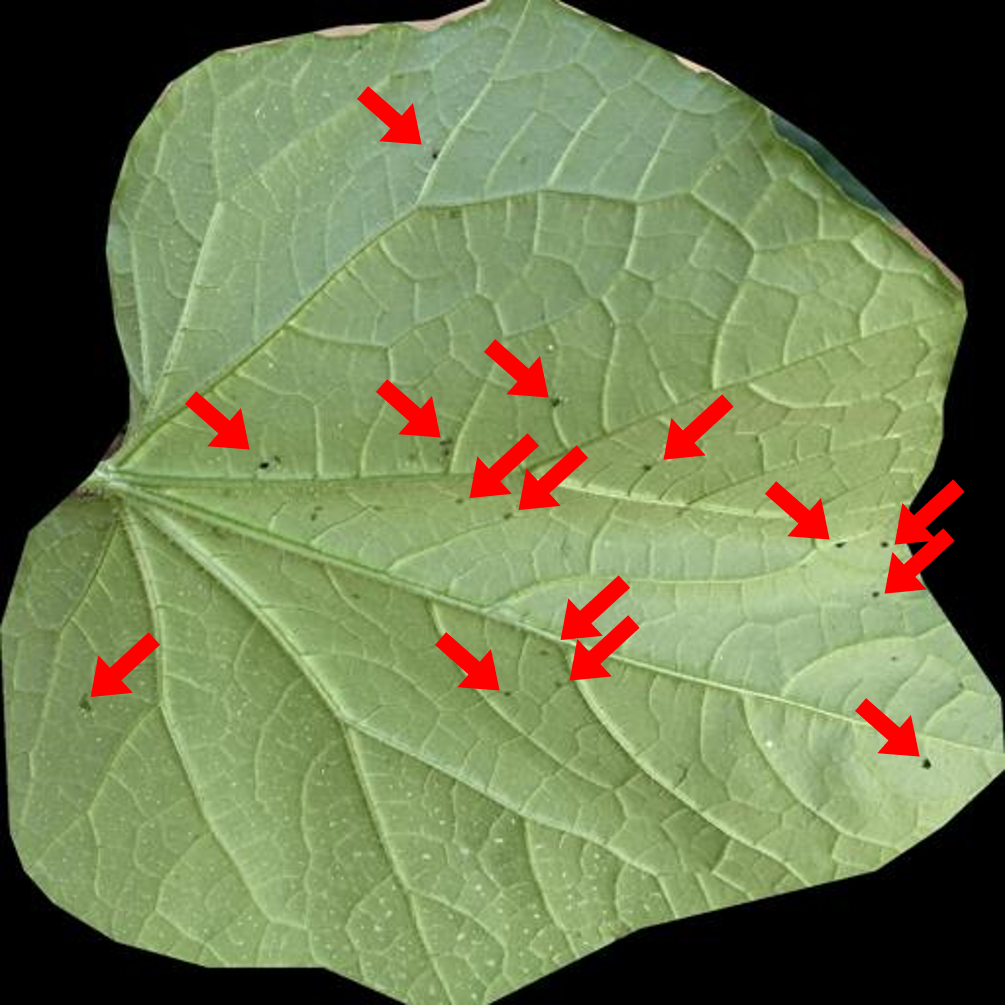}}
 \subfigure{\includegraphics[]{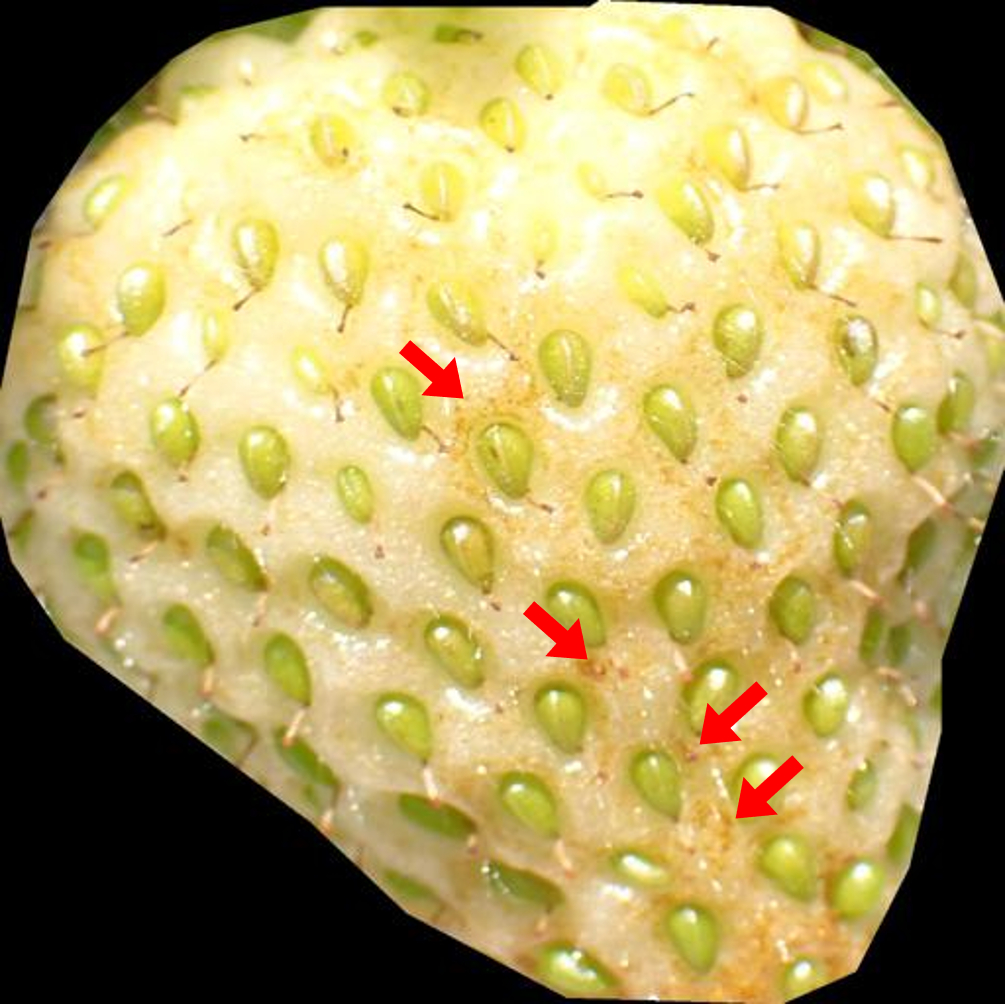}}
 \subfigure[Whitefly (6)]{\includegraphics[]{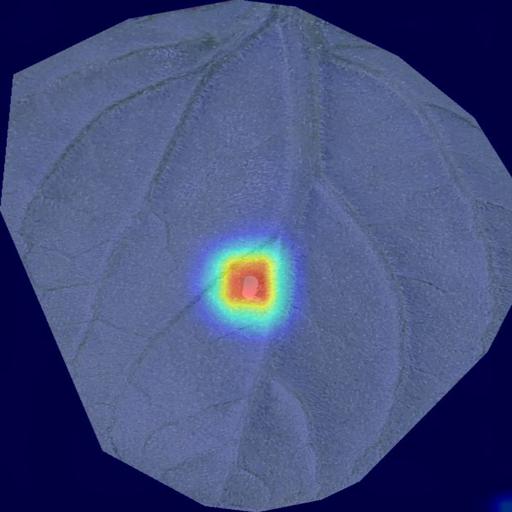}}
 \subfigure[Aphid (7)]{\includegraphics[]{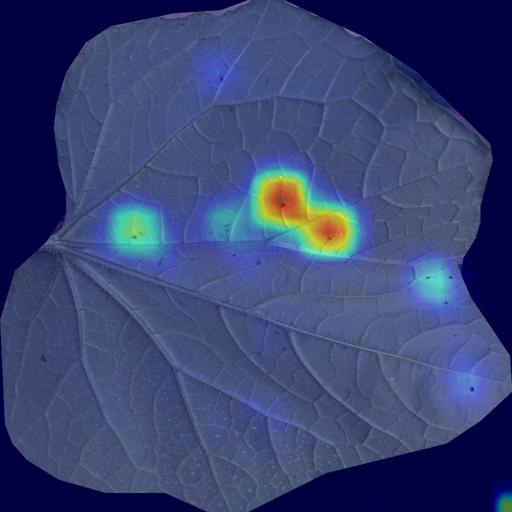}}
 \subfigure[Thrips (10)]{\includegraphics[]{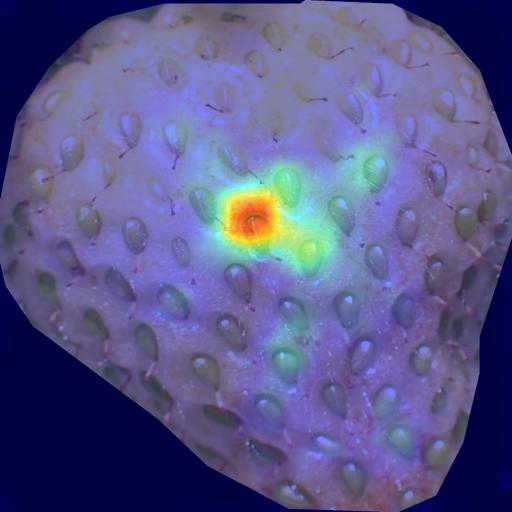}}
\end{subfigmatrix}
\caption{Examples of accurately identified pest species. Upper row: original image, middle row: images after ROI extraction, lower row: Grad-CAM. Arrows indicate the position of the insect body in (a) and (b) and the slight browning caused by feeding damage in (c).}
\label{fig:fig8}
\end{figure}

\begin{figure*}
\setcounter{subfigure}{-12} 
\begin{subfigmatrix}{6}
 \subfigure{\includegraphics[]{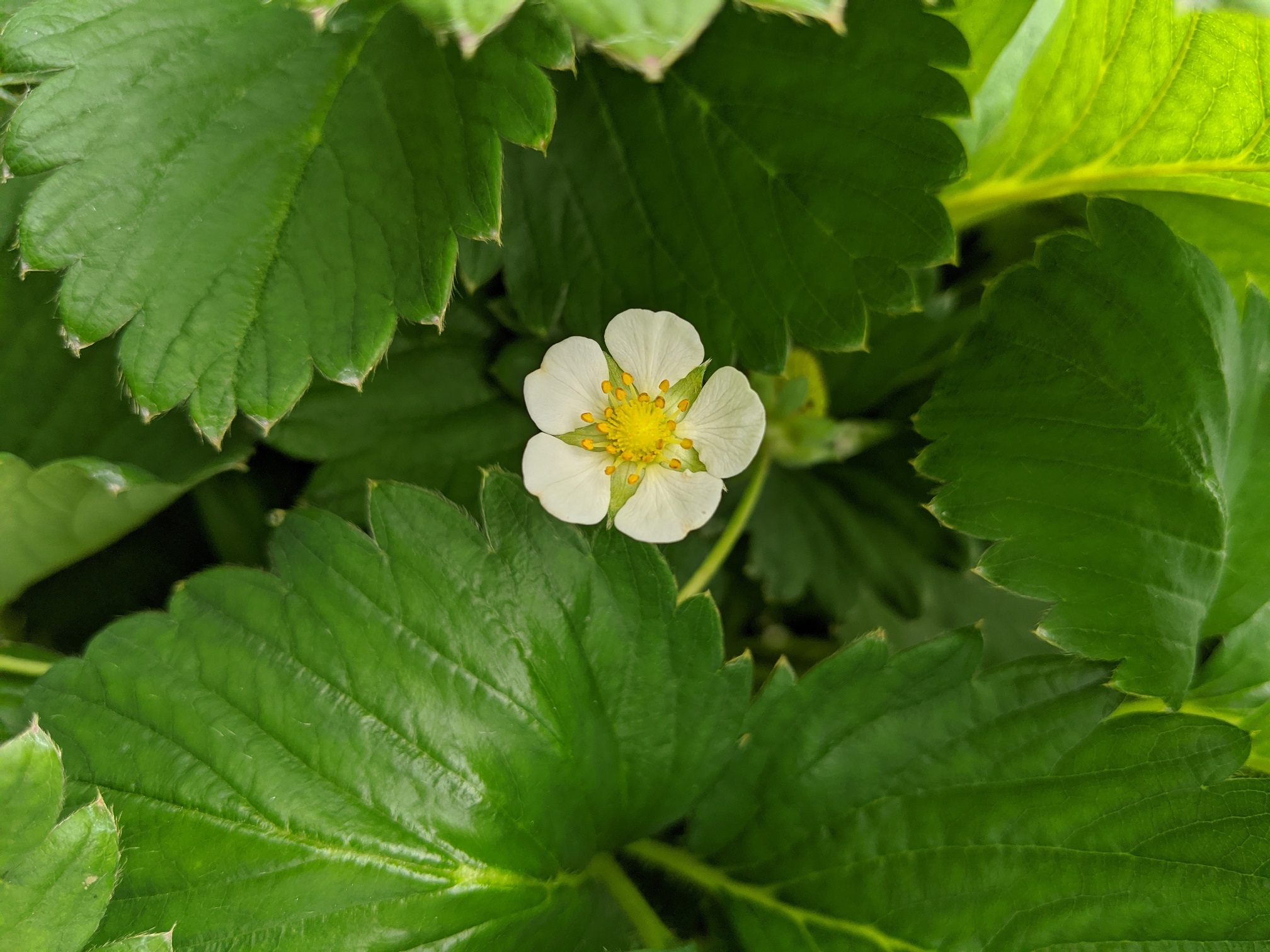}}
 \subfigure{\includegraphics[]{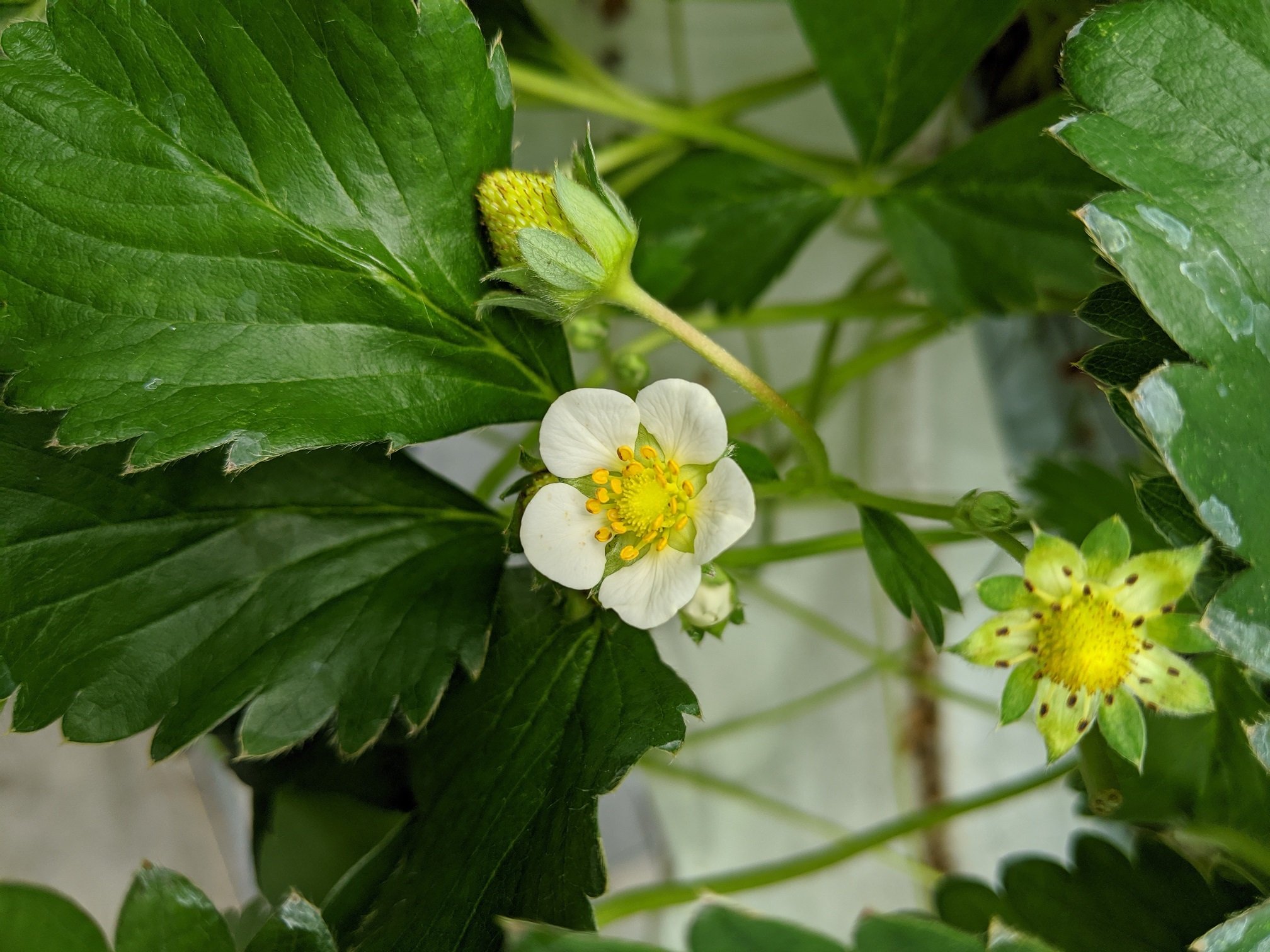}}
 \subfigure{\includegraphics[]{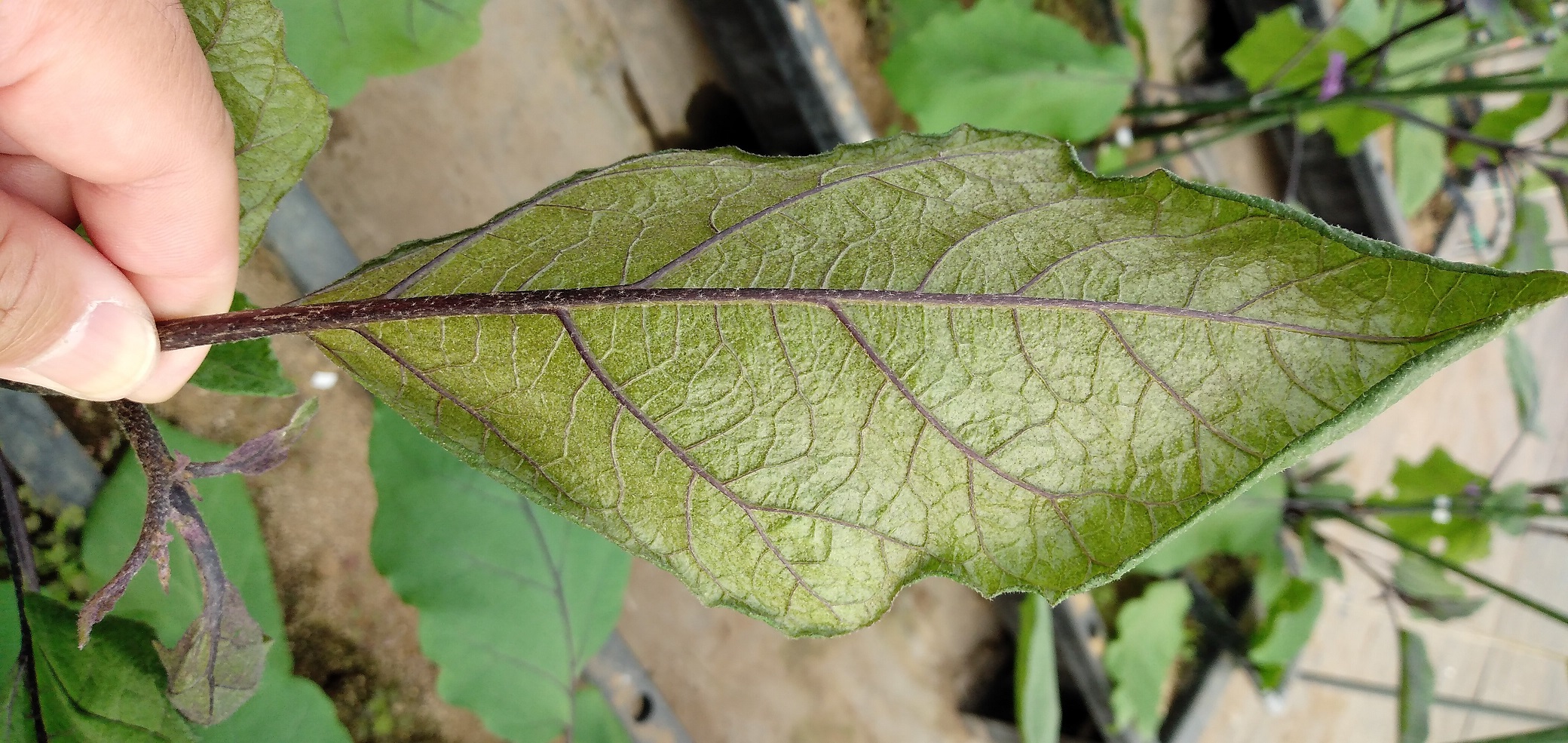}}
 \subfigure{\includegraphics[]{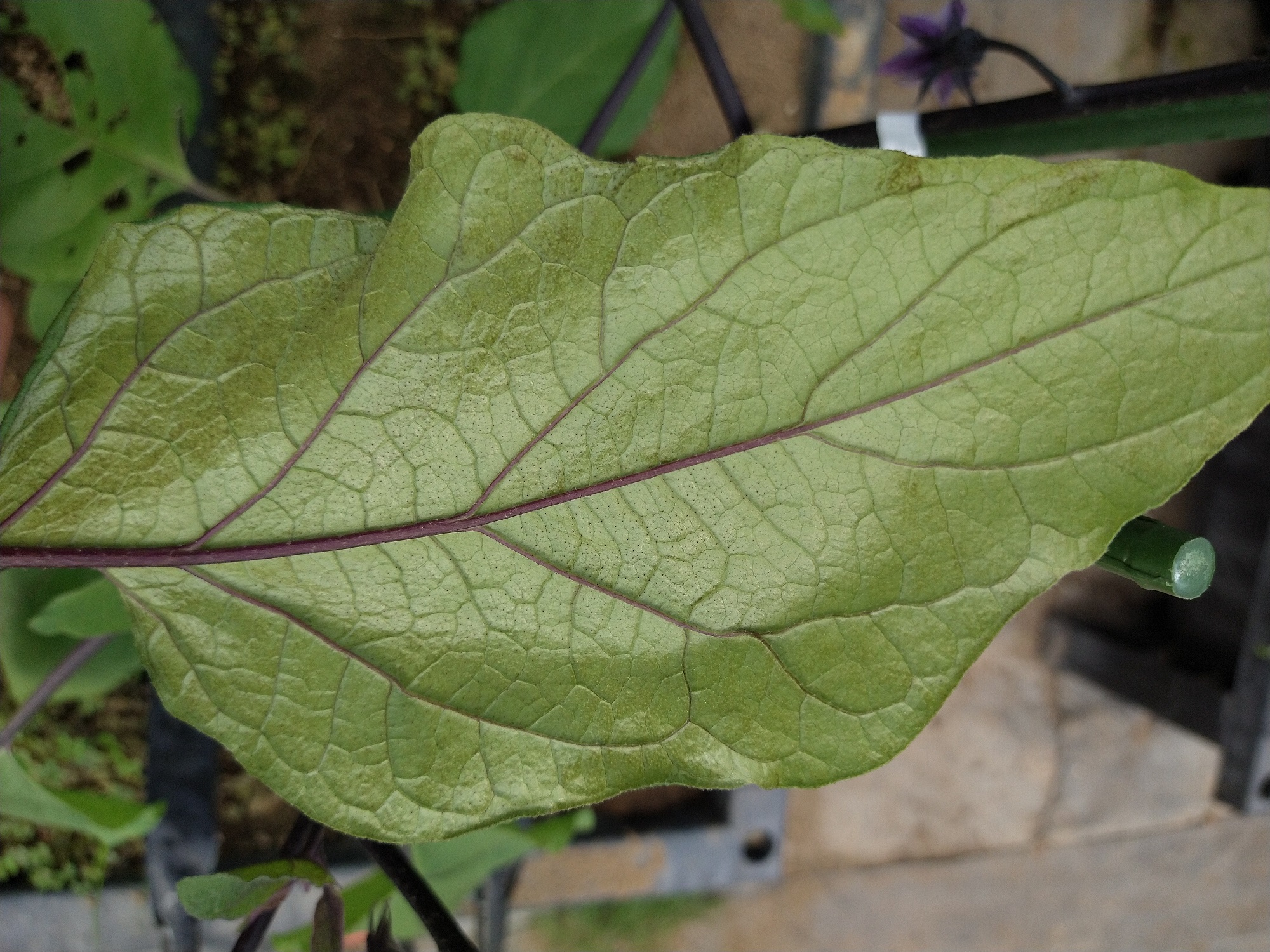}}
 \subfigure{\includegraphics[]{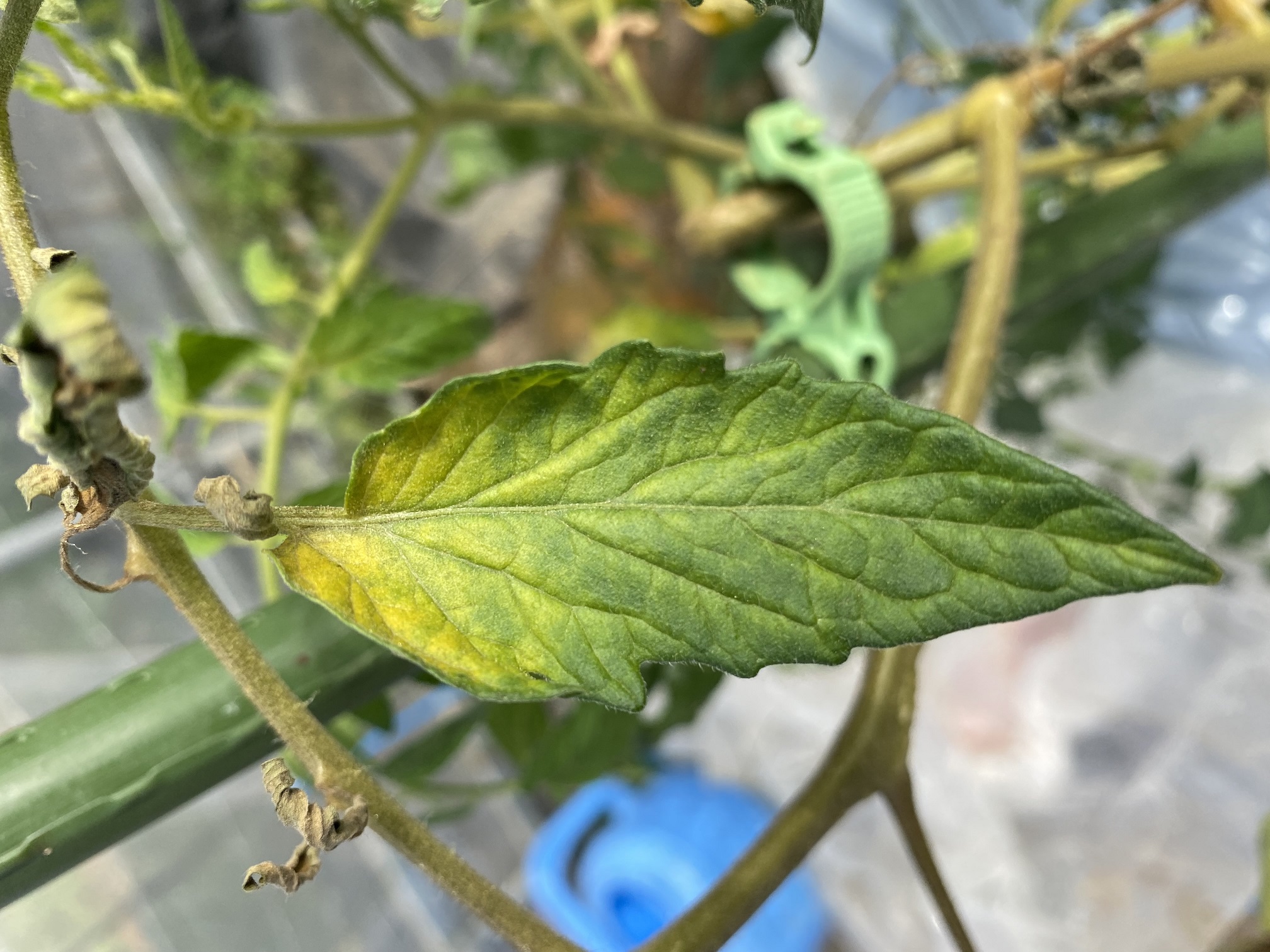}}
 \subfigure{\includegraphics[]{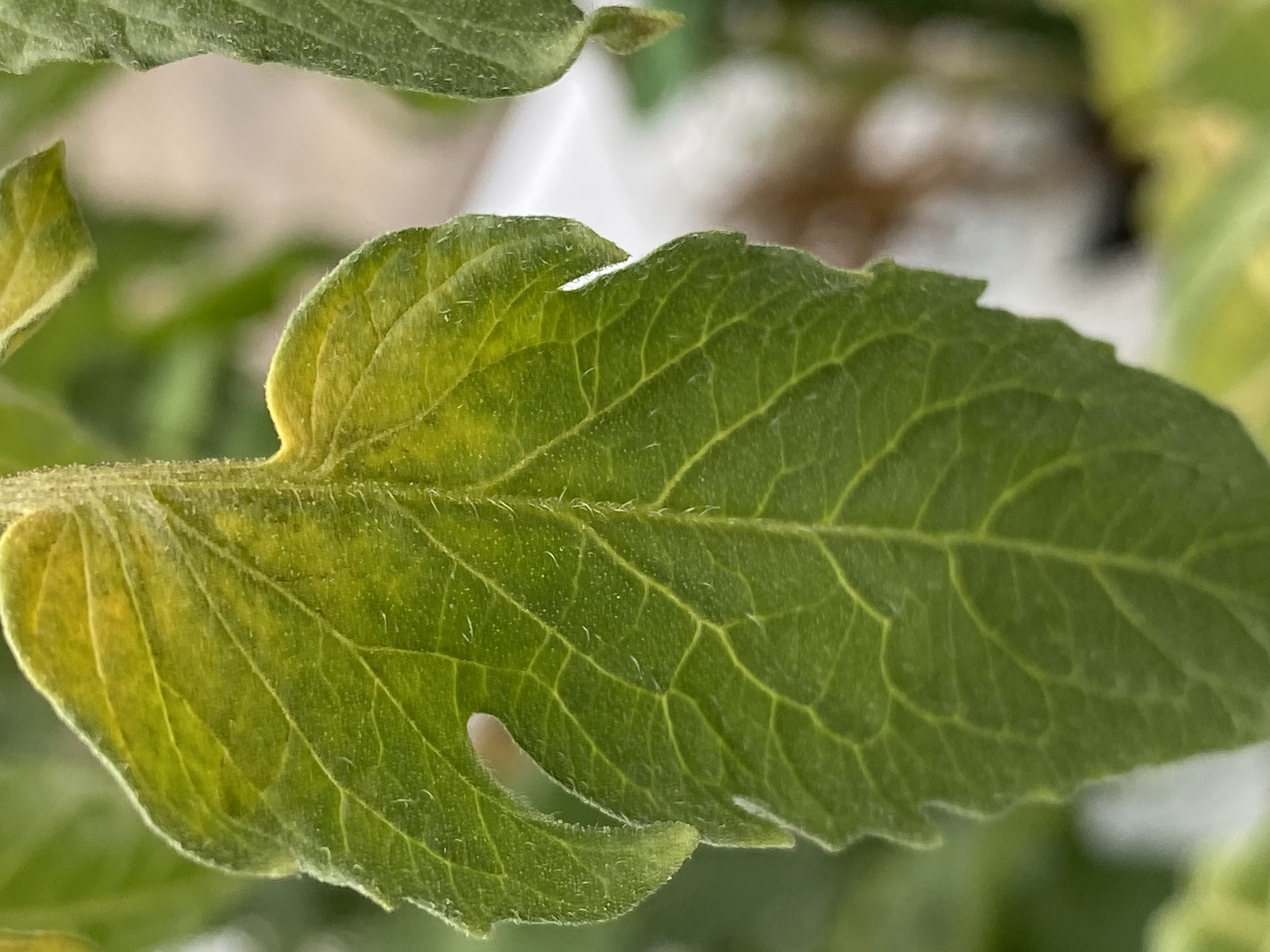}}
 \subfigure{\includegraphics[]{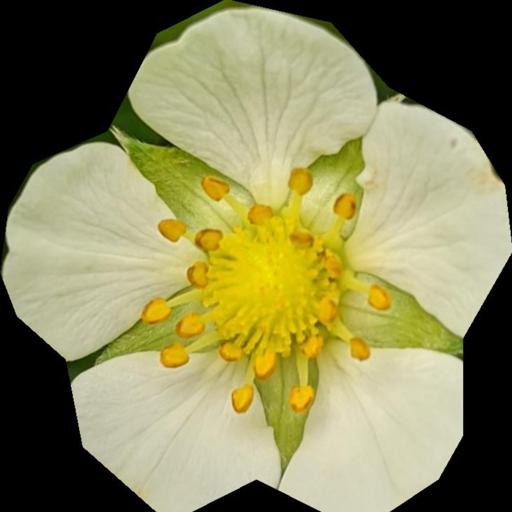}}
 \subfigure{\includegraphics[]{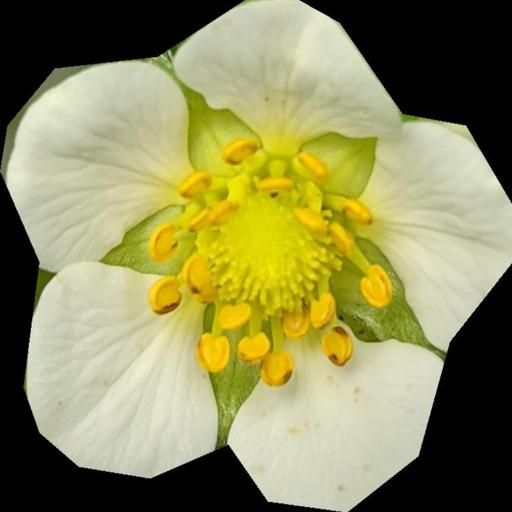}}
 \subfigure{\includegraphics[]{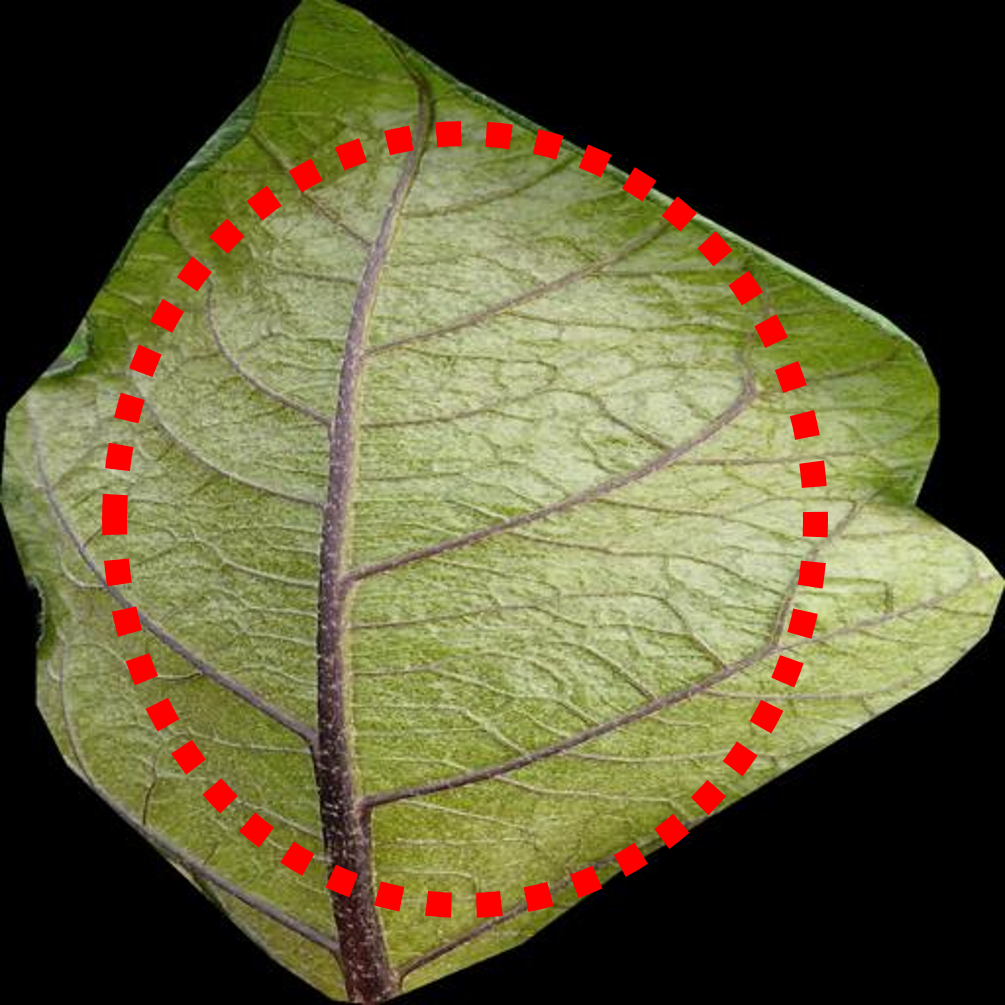}}
 \subfigure{\includegraphics[]{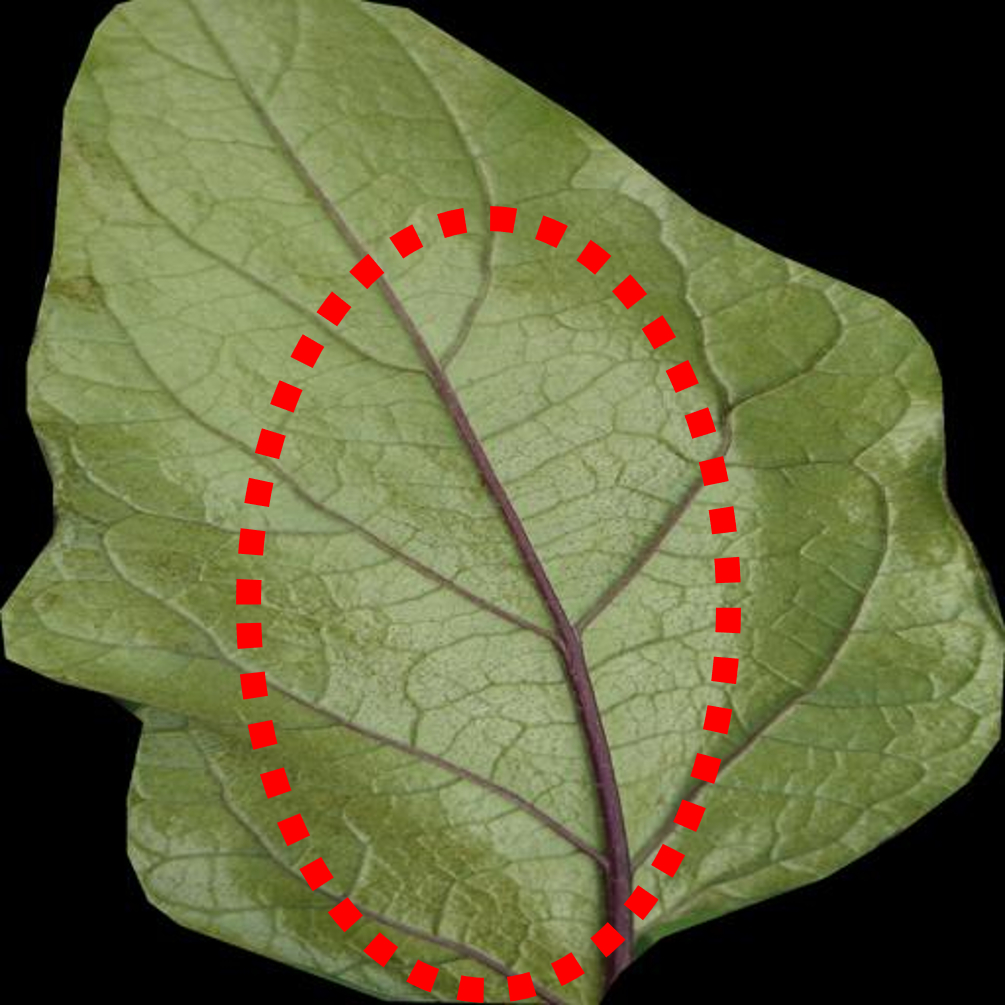}}
 \subfigure{\includegraphics[]{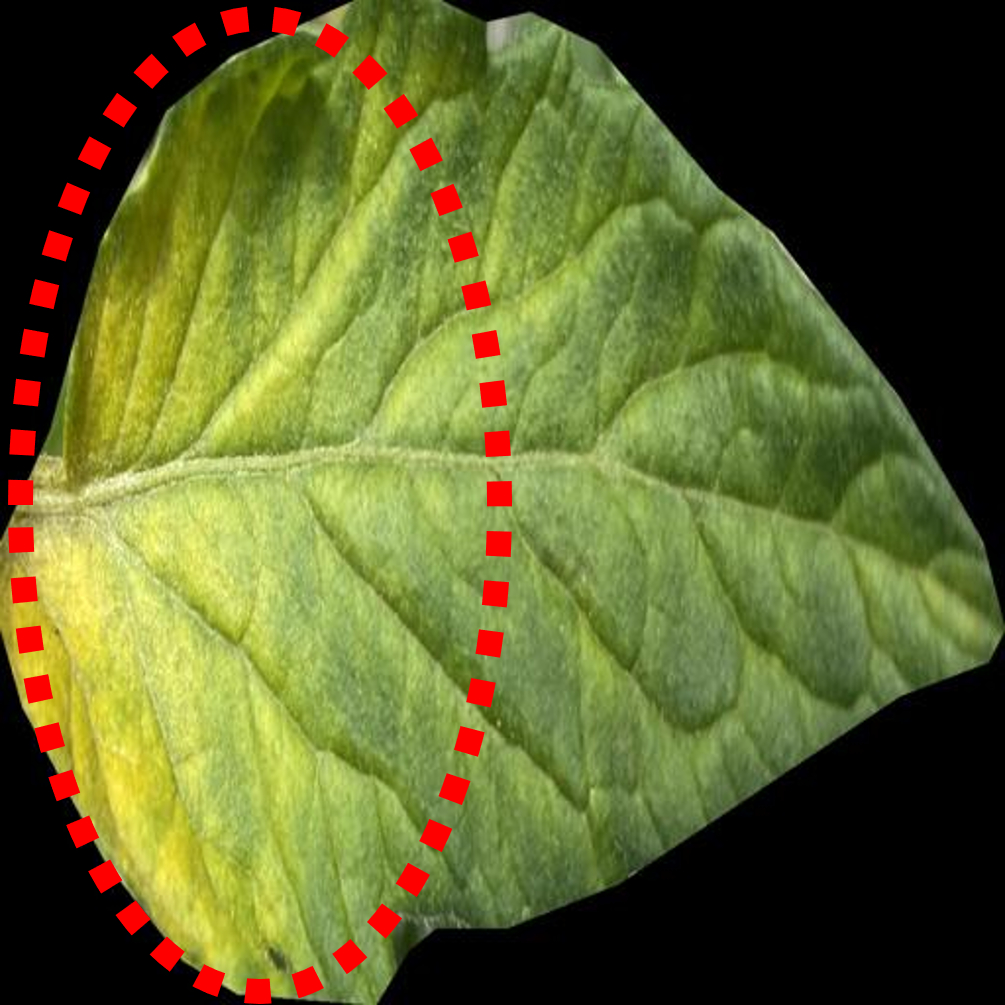}}
 \subfigure{\includegraphics[]{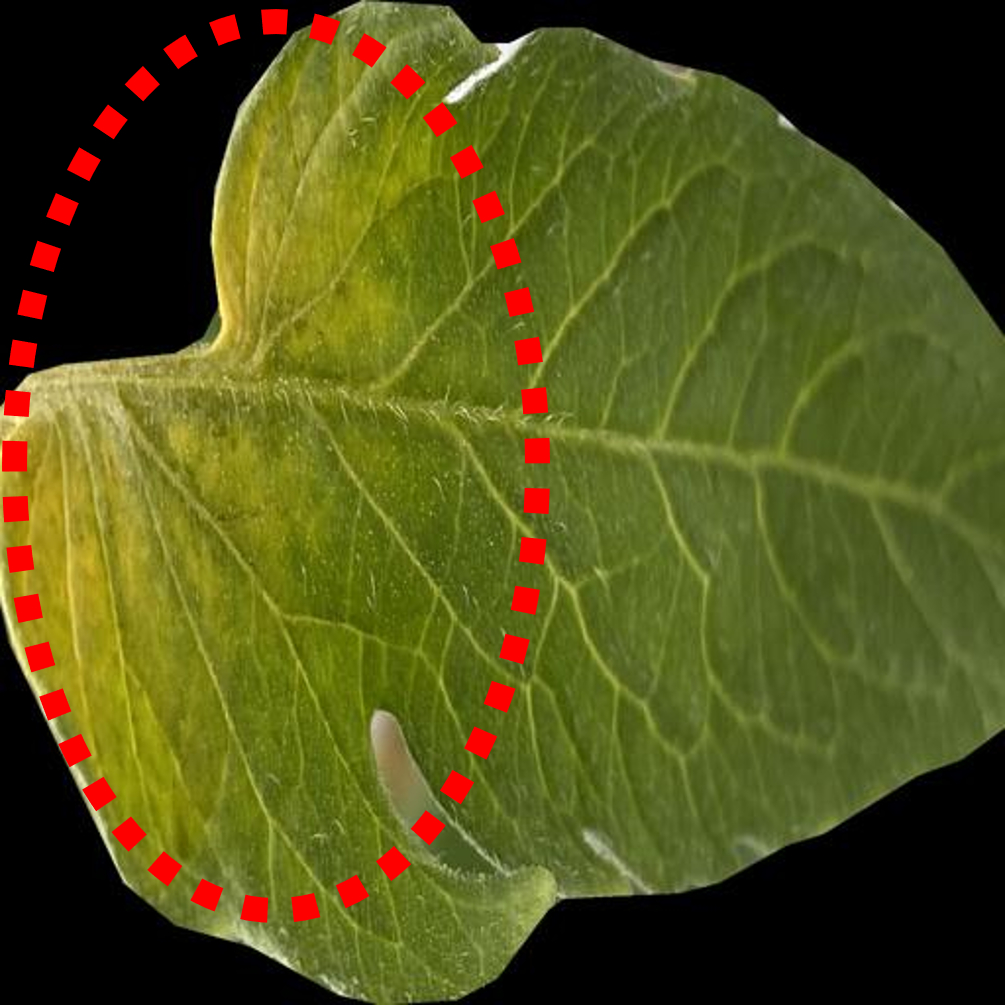}}
 \subfigure[Correct]{\includegraphics[]{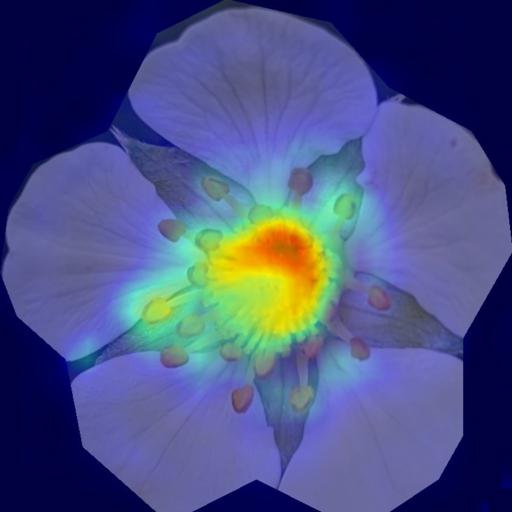}}
 \subfigure[Misidentified as thrips \newline (Fig.7(i): (23)$\rightarrow$(12))]{\includegraphics[]{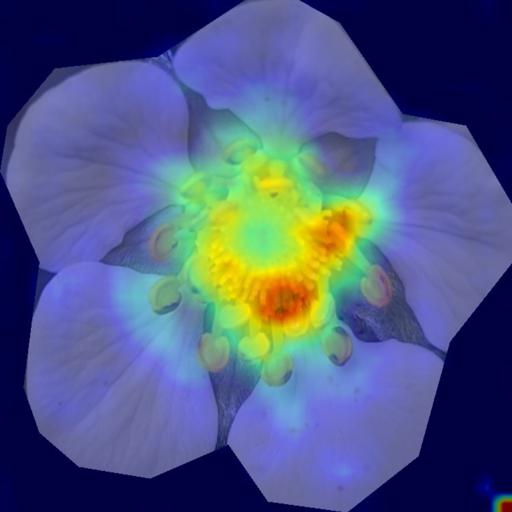}}
 \subfigure[Correct]{\includegraphics[]{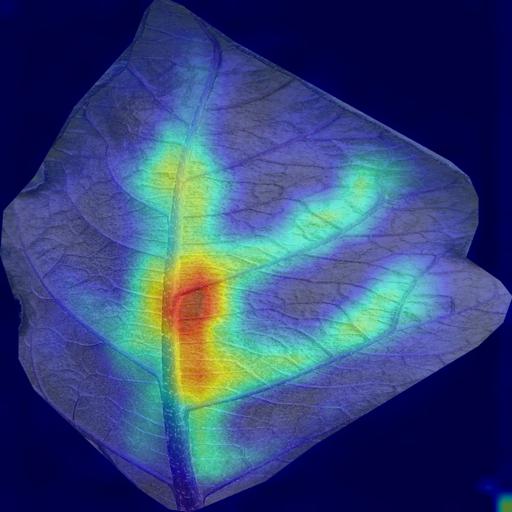}}
 \subfigure[Misidentified as healthy \newline (Fig.7(ii): (3)$\rightarrow$(25))]{\includegraphics[]{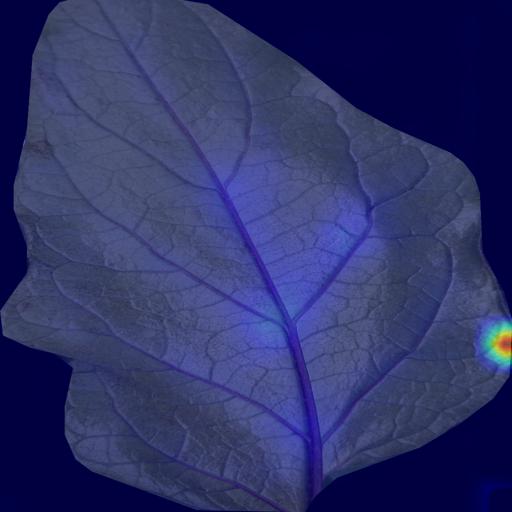}}
 \subfigure[Correct]{\includegraphics[]{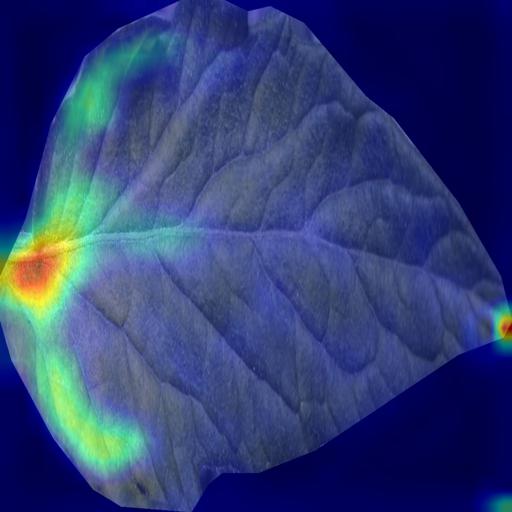}}
 \subfigure[Misidentified as healthy \newline (Fig.7(iii): (5)$\rightarrow$(25))]{\includegraphics[]{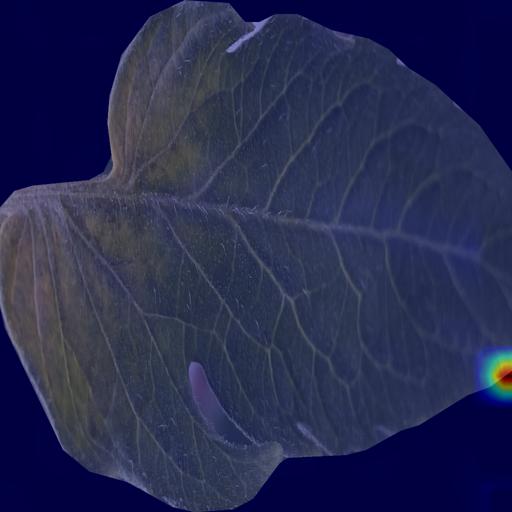}}
\end{subfigmatrix}
\begin{minipage}[t]{0.30\linewidth}
\caption*{(I) Healthy (strawberry flower) (23)}
\end{minipage}
\hfill
\begin{minipage}[t]{0.30\linewidth}
\caption*{(II) Broad mite (cucumber leaf) (3)}
\end{minipage}
\hfill
\begin{minipage}[t]{0.30\linewidth}
\caption*{(III) Tomato russet mite (5)}
\end{minipage}
\hfill
\caption{Comparison of successful and unsuccessful identification examples for low identification categories. \\ {\footnotesize Upper row: Image before ROI extraction, middle row: Image after ROI extraction, lower row: Grad-CAM. The areas the experts indicated as the basis for identification are indicated by dashed lines.}}
\label{fig:fig9}
\end{figure*}

\section{Discussion}
\subsection{Impact of training and test data separation}
%
%
In RQ1, we investigated the effect of data partitioning criteria on measured model performance, based on the hypothesis that images from the same farm field are likely to have common features that may affect the generalizability of a model. 
The result revealed a significant degradation in the classification performance for the different farm scenario compared to the same farm scenario, despite a relative uniformity in composition and lighting conditions among the images.
The results of experiment RQ1 extend the applicability of insights from plant disease diagnosis research \cite{Saikawa2019, Shibuya2021, Suwa2019, Mohanty2016, Ferentinos2018, Cap2020, Kanno2020, Guth2023, Wu2023, Xu2023}.
The susceptibility of a discriminator to irrelevant features may be exaggerated by a disproportional representation of the target features (pests, diseases) and irrelevant attributes (image brightness, composition, background etc.) in an image -- as the target features are often small or indistinct, the model may be inclined to learn more visible features, easily overfitting the dataset.
At a fundamental level, the primary cause of deterioration of a model in unseen contexts is attributed to a lack of diversity in the training data, thereby suggesting that the problem can be overcome by adding a sufficient amount of diverse training data. 
However, this approach may not be practical \cite{Xu2023}, as the cost associated with data collection and cleansing would also increase. 
The fact that the training dataset used in RQ1 has 9K images, which is comparable to some of the largest datasets used in previous studies, further undermines its feasibility.
%

%
On the other hand, in scenarios with limited training data, enhancing image diversity is crucial for improving model performance.
Recent research highlights the efficacy of using generative models like GANs \cite{Cap2020, Kanno2020} or recent latent diffusion models for data augmentation and generation.
Especially the latter can generate high quality images from a small dataset and can generate a variety of images according to textual instructions, showing great promise by utilizing the huge amount of prior learning information. Another effective method is to apply the domain adaptation technique, which harmonizes domain differences such as field differences, to low-dimensional representations of images \cite{Wu2023}.
However, rigorous assessment of model performance post-application of these techniques remains essential.

In conclusion, we believe that rigorous separation of data sources is the most viable strategy for discerning the discrepancy between the in- and out-of-experiment performances, as exemplified in the “different farm” scenario. 
The exact methodology for partitioning may vary depending on the context of model development---the location of image capture, the source dataset, or other attributes may be used as a proxy for potential shared features, provided that the images are collected in different contexts. 
Further studies are warranted for quantitative assessment of potential factors that may affect the generalizability of discriminators.
%

\subsection{Impact of ROI detection}
%
%
The results for RQ2 and the development of the main model (Table 2, Figures 5, 8, and 9) reveal that ROI pre-extraction succeeds in magnifying small cues of pest damage and insect bodies and effectively directs the model’s attention, thereby that background removal is an efficient method to improve pest identification performance.
CNNs, which are commonly used as classification models, are inherently sensitive to the size of the object in the image and the distance to the object; thus data augmentation techniques such as random cropping are commonly used, but they are insufficient for the identification of very fine-grained pest damage.
Object detection models overcome these limitations of CNNs and have the ability to detect very small objects, but, as mentioned above, the training cost is high.
Particularly in the case of pest damage, there are numerous cases in which it is difficult to annotate not only the body of the insect, but also features unrelated to the insect, such as feeding scars and invisible microscopic insects, as clues for diagnosis.
This makes ROI detection an effective tool for pest damage diagnosis due to its low learning and execution costs.

In contrast, as reported by Shibuya et al. \cite{Shibuya2021}, in plant disease diagnosis, the performance improvement of ROI detection diminishes when the training image dataset is large and diverse and the image resolution is high (512 $\times$ 512 pixels).
This difference in the effectiveness of ROI detection can be attributed to two major differences between plant diseases and pests.
First, visual signs of plant diseases (lesions) generally appear larger in the image than those of pests (insect bodies, feeding scars, etc.). 
Second, lesions are often more varied in appearance than pest damage.
In other words, the two tasks are similar in terms of the difficulty of making an identification based on small, faint clues, but they differ in terms of the quality of the difficulty.

In summary, it is evident that the pre-detection of ROI is effective for pest damage, even if it is not as effective for disease.
The intended users of the plant diagnostic system we are developing are primarily those with little expertise in the field and those who may have difficulty distinguishing between diseases and pests.
A practical system should target both, and a mechanism to pre-detect ROIs for input images is important.
On the other hand, even if ROI detection can further reduce the influence of external factors at the time of image acquisition, there is still room for improvement in discrimination accuracy. Further studies in this are expected in the future.

\subsection{Impact of class definition}
%
In a comparative experiment to answer RQ3, improved performance was observed in both the integrated scenario and the cross-crop scenario.
The integration of closely related species performed in this study was based on selecting species for integration that are very similar in appearance, as depicted in Figure 4, and considered identical in the pest control process.
In the baseline model, much confusion was observed within each of the spider mite and aphid classes, but the integration trick reduced the number of false positives and greatly improved performance. However, it is important to consider issues such as similarity between classes and the operational goals of the model when constructing the dataset.

Although the addition of images of crops other than the target plants raises bias concerns due to class imbalance \cite{Wu2019}, this experiment revealed that the benefits of diversity outweighed the consequences of bias due to improved overall model performance.
Our study also confirmed that adding only healthy plant images of other crops improved performance to a certain extent. 
Because images of healthy plants are more readily available than images of pest-damaged samples, this strategy deserves attention as a simple but effective means to improve identification performance.
However, it should be noted that some pest categories may exhibit different damage trails from crop to crop; thus, this method is only feasible when visual characteristics are independent of the crop type, such as when insect bodies and excrement are clues of pest detection.
Furthermore, our experiments confirmed that the simultaneous application of the abovementioned two techniques further improved the performance. 
Thus, we expect that the methods proposed to answer this RQ will be applicable when robust models are built with datasets that contain multiple crops and closely related pest species.

\subsection{Discussion of the main results}
The identification framework, using each of the methods investigated in each RQ, achieved an accuracy and macro F1 score of approximately 90\%, and per-class F1 scores greater than 80\% in 18 of the 21 classes. Unlike many of the previous studies, the performance of this model in real fields is expected to be comparable to these results, as the model was trained and evaluated with strictly separated agricultural fields (see RQ1).

Figure 7 indicates that most of the detection results are on the diagonal line of the confusion matrix, thereby indicating that the model has successfully learned the pest features.
The figure also demonstrates the effectiveness of the class integration introduced in RQ3, as most of the failure cases were observed between a pest class and the healthy class of the corresponding crop and portion and not between pest species.
The effect of ROI extraction (see RQ2) is evident in Figure 8.
The figure depicts correctly identified cases of (6) whiteflies, (7) aphids, and the browning caused by (10) thrips, all of which have small or obscure class features (insect bodies and damage). 
As depicted in the first row of Figure 8, these features are barely visible in the original images.
ROI detection works as a magnifier by trimming unnecessary portions of images, as well as omitting background information. 
Consequently, the model directed its attention to pest features, thereby leading to accurate classification results.

However, the result also reveals the limitations of the model's diagnostic performance. 
Figure 9 depicts a comparison of successful and failed cases in three low-performing classes.
In the cases of healthy strawberry flowers (23), the model focused on the same area (stamen and pistil at the center of the flower), but it yielded different discrimination results (“healthy" and “thrips").
This over-detection of thrips was caused by instances of the pest that are extremely difficult to distinguish; Figure 10 indicates that, in certain images of thrips infesting flowers, the model paid attention to the stamens and pistil instead of the insect body. While the ROI extraction is effective against small insect bodies on leaves, these examples illustrate the difficulties of identifying them on plant body portions with relatively complex visual features like flowers.
The model also failed to identify damages of broad mite (3) and tomato russet mite (5), both of which are characterized by texture changes (silvering and browning).
Such characteristics pose different challenges in distinguishing small features, since a variety of factors may affect how they may appear in images, such as weather, time of day, and the angle of the camera.
These cases reveal that, while the methods introduced in RQ2 and RQ3 may help models in identifying small pests, they may be less effective in identifying pest damages characterized by changes in texture or color of plants, or for early detection of pests infesting on certain portions of plant bodies.
Meanwhile, another instance of a class with low performance, healthy strawberry fruits (19), is apparently caused by data bias, as the class has a disproportionally small number of images (n = 27) compared to others. 
This is a common phenomenon observed in ML, and the model’s “real” performance on this class is assumed to be comparable, as the recall of this class is high (recall = 100).

Overall, the result proves the effectiveness of the methods discussed in three RQs in a scenario that employs a large-scale dataset. 
%
Thanks to the results from RQ2 and RQ3, pest damage can be identified with high accuracy using only images taken with a standard camera, without the need for magnified images of insect bodies as in previous studies. 
This shows that the framework is not only user-friendly, but also promises a wider application than those requiring dedicated equipment, since our approach can handle cases where insect bodies are not visible.
%
While the model showed its weaknesses in identifying certain pest classes characterized by textural changes, they may be mitigated by building a separate discriminator specialized for classifying such features. Further investigations are required in this area.

\setcounter{subfigure}{-2} 
\begin{figure}
\begin{subfigmatrix}{2}
 \subfigure{\includegraphics[width=0.25\textwidth]{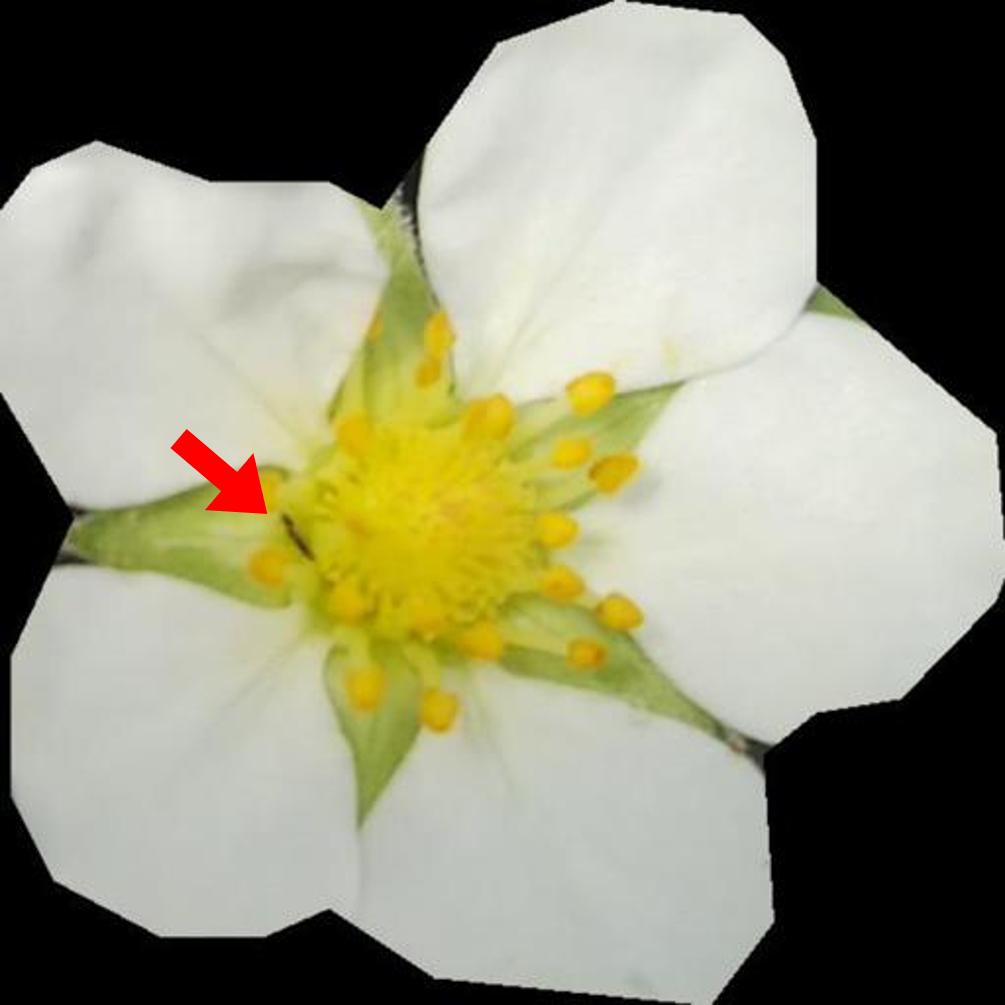}}\hfill
 \subfigure{\includegraphics[width=0.25\textwidth]{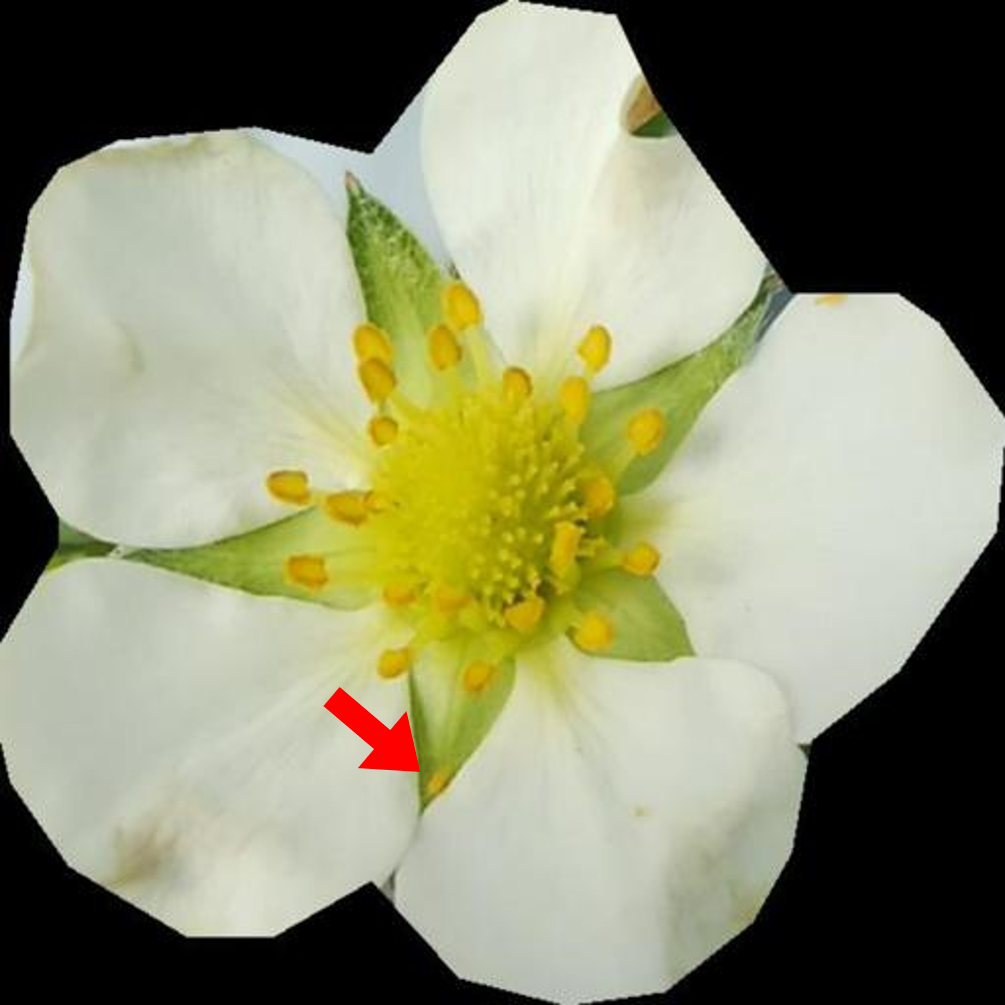}}
 \subfigure[Focused on the insect body]{\includegraphics[width=0.25\textwidth]{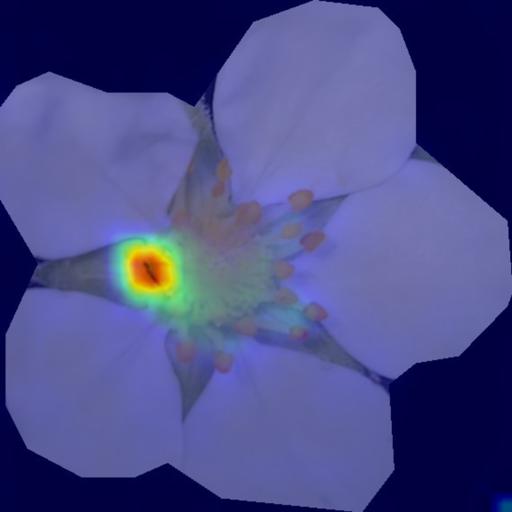}}\hfill
 \subfigure[Focused on the stamen and pistil]{\includegraphics[width=0.25\textwidth]{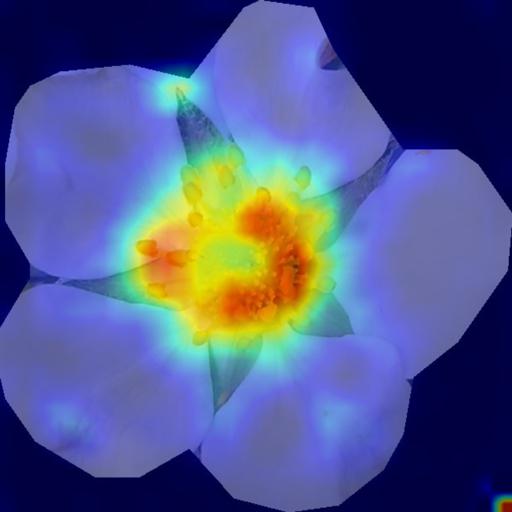}}
\end{subfigmatrix}
\caption{Differences in Grad-CAM's attention to thrips (flower strawberry) (12). \\{\footnotesize Arrows indicate the location of the insect. Both images are identified correctly, but (a) is focused on the insect body and (b) on the stamen and pistil, respectively.}}
\label{fig:fig10}
\end{figure}

\subsection{Study limitations}
%
%
The study has revealed four areas that warrant further investigation.
(1) The study has not revealed which external factor may contribute to the superficial improvement of model performance. Instead, we assumed data sources (fields of image capture) to be a collective representation of the irrelevant features. 
(2) The study targeted the problem of identifying only one type of pest damage per image, not multiple types of pest damage and disease co-infestation.
(3) In the proposed two-stage framework, when the outer edge of a leaf or fruit is severely damaged, in the first stage, the foreground extraction model recognizes the damaged area as the background and eliminates it; in the second stage, the model submits an incorrect diagnosis, although only in rare cases.
(4) With regard to the performance improvement in class merging and supplementing images of other crops, as indicated in RQ3, the effect has currently only been confirmed for leaf images, which are the most widely studied, but has not been evaluated for fruit or flower images.

Issue (1) remains an open question, and further research is needed for accurate measurement of model generalizability. For example, analyses of model susceptibility to differences in image contexts (e.g. lighting, composition) may yield crucial insights on data separation strategies. 
Issue (2) can be addressed by extending the diagnostic model to multi-label or, in the case of simultaneous disease and insect infestation, by combining disease diagnostic models.
A supplementary solution to issue (3) is an alternative model that operates parallelly with the proposed two-stage framework. 
This model would retain the image background and then crop portions of leaves, fruits, and flowers before employing a CNN to identify pest damage, thereby mitigating the shortcomings of background elimination.

\section{Conclusion}
In this study, we investigated three key considerations for the development of practical pest identification models. A comparative experiment (RQ1) illustrated the importance of data separation for assessing generalizability of a discriminator by revealing a significant performance degradation when rigorous data separation of data sources was performed between training and test sets. 
This study also investigated three strategies for improving pest discrimination performance---ROI extraction (RQ2), class integration, and cross-crop training (RQ3)---and demonstrated their efficacy through experiments in various scenarios. 
Our two-stage pest identification framework, consisting of ROI extraction and identification phases, exhibited an extremely high identification performance, with an average accuracy and macro F1-score of 91.0\% and 88.5\%, respectively, even though the model was used to evaluate images from fields that were not seen during training. The dataset used in this study is available to the public through the National Institute of Agrobiological Sciences in Japan. We hope our research will contribute to the advancement of plant pest and disease research.

%
\appendix
\section{Scientific names}
Table A summarizes the correspondence between the common names of pest species analyzed in this study and their scientific names.
Certain species are integrated because they are close relatives and share the same extermination methods.
For more information, see RQ3 and the main experiment sections.

\renewcommand{\thetable}{\Alph{section}}
\begin{table}[t]
\begin{center}
\caption{Correspondence between the generic and scientific names of the pests used in this experiment}
\small
\begin{tabular}{ r l l l} \hline
ID$\dag$  & name$\dag$ & common name & scientific name \\ \hline
1,2,3 & broad mite & broad mite & {\it Polyphagotarsonemus latus} (Banks) \\
4     & spider mite & Kanzawa spider mite & {\it Tetranychus kanzawai} Kishida \\
      &  & twospotted spider mite & {\it Tetranychus urticae} Koch \\
5 & tomato russet mite & tomato russet mite & {\it Aculops lycopersici} (Massee) \\
6 & whitefly & greenhouse whitefly & {\it Trialeurodes vaporariorum} (Westwood) \\
 &  & tobacco whitefly & {\it Bemisia tabaci} (Gennadius) \\
7 & aphid & cotton aphid & {\it Aphis gossypii} Glover \\
 &  & green peach aphid & {\it Myzus persicae} (Sulzer) \\
8,9 & mealybug & Solanum mealybug & {\it Phenacoccus solani} Ferris \\
 &  & Madeira mealybug & {\it Phenacoccus madeirensis} Green \\
10,11,12,13 & thrips & western flower thrips & {\it Frankliniella occidentalis} (Pergande) \\
 &  & melon thrips & {\it Thrips palmi} Karny \\
 &  & onion thrips & {\it Thrips tabaci} Lindeman \\
 &  & Eurasian flower thrips & {\it Frankliniella intonsa} (Trybom) \\
14 & hadda beetle & hadda beetle & {\it Henosepilachna vigintioctopunctata} (Fabricius) \\
15 & leaf miner & vegetable leafminer & {\it Liriomyza sativae} Blanchard \\
 &  & serpentine leafminer & {\it Liriomyza trifolii} (Burgess) \\
 &  & tomato leafminer & {\it Liriomyza bryoniae} (Kalgenbach) \\
16,17 & cotton bollworm & cotton bollworm & {\it Helicoverpa armigera} (Hübner) \\
18 & tobacco cutworm & tobacco cutworm & {\it Spodoptera litura} (Fabricius) \\ \hline
\end{tabular}
\end{center}
{\small $\dag$: ID and (integrated) name used in the main experiment, corresponding to Table 5. \\}
\end{table}

 \section*{Acknowledgment}
This work was supported by the Japanese Ministry of Agriculture, Forestry and Fisheries (MAFF) commissioned project study on the development of pest diagnosis technology using AI (JP17935051) and by the Cabinet Office, Public/Private R\&D Investment Strategic Expansion PrograM (PRISM).
We would like to express our sincere thanks to all the experts and test site personnel who actually grew the plants in their respective fields, strictly controlled pests and diseases, and took an extremely large number of high quality photographs in conducting this study.
We also would like to express our sincere gratitude to Dr. Junsuke Yamasako for his valuable advice guidance in the preparation of the pest species list.

\bibliographystyle{IEEEtran}
\bibliography{main}

\end{document}